\documentclass[letterpaper]{article} % DO NOT CHANGE THIS
\usepackage[submission]{aaai2027}  % DO NOT CHANGE THIS
\usepackage[hyphens]{url}  % DO NOT CHANGE THIS
\usepackage{graphicx} % DO NOT CHANGE THIS
\usepackage{natbib}  % DO NOT CHANGE THIS AND DO NOT ADD ANY OPTIONS TO IT
\usepackage{caption} % DO NOT CHANGE THIS AND DO NOT ADD ANY OPTIONS TO IT
\usepackage{algorithm}
\usepackage{algorithmic}

\usepackage{newfloat}
\usepackage{listings}
\DeclareCaptionStyle{ruled}{labelfont=normalfont,labelsep=colon,strut=off} % DO NOT CHANGE THIS
\floatstyle{ruled}
\newfloat{listing}{tb}{lst}{}
\floatname{listing}{Listing}

\usepackage{booktabs}

\usepackage{amssymb}
\usepackage{amsthm}
\usepackage{amsmath,amsfonts,bm}

\def\eqref#1{equation~\ref{#1}}
\def\floor#1{\lfloor #1 \rfloor}
\def\1{\bm{1}}

\def\rvq{{\mathbf{q}}}

\DeclareMathAlphabet{\mathsfit}{\encodingdefault}{\sfdefault}{m}{sl}
\SetMathAlphabet{\mathsfit}{bold}{\encodingdefault}{\sfdefault}{bx}{n}

\newcommand{\E}{\mathbb{E}}

\newcommand{\R}{\mathbb{R}}

\newcommand{\softmax}{\mathrm{softmax}}

\DeclareMathOperator*{\argmax}{arg\,max}

\newcommand\abs[1]{\vert#1\vert}

\newcommand\norm[1]{\Vert#1\Vert}

\newtheorem{theorem}{Theorem}

\usepackage{multirow}
\nocopyright %-- Your paper will not be published if you use this command
\title{AnchorKV: Anchor-Residual KV Cache Compression}
\author {
    Malik Khalaf\textsuperscript{\rm 1}\equalcontrib,
    Yara Shamshoum\textsuperscript{\rm 1}\equalcontrib,
    Nitzan Hodos\textsuperscript{\rm 1},
    Yuval Sieradzki\textsuperscript{\rm 1},
    Assaf Schuster\textsuperscript{\rm 1},
}
\affiliations {
    \textsuperscript{\rm 1}Department of Computer Science, Technion - Israel Institute of Technology\\
    \{malikkhalaf, yara-sh, hodosnitzan, syuvsier\}@campus.technion.ac.il
}

\begin{document}

\maketitle
% Added for preprint at the bottom
\let\thefootnote\relax
\footnotetext{Preprint. Under review.}
\footnotetext{Our code will be released upon acceptance.}
% then after \maketitle:
% \thispagestyle{firstpage}
%%%%%
\begin{abstract}
The key-value (KV) cache is the primary memory bottleneck in long-context LLM inference. Existing approaches attack it from opposite ends: eviction methods permanently discard tokens, degrading performance whenever a discarded token later proves essential, while quantization methods retain all tokens at low precision but offer limited compression. We propose AnchorKV, a compression scheme that shrinks the cache by 20$\times$ without discarding a single token. AnchorKV represents the cache using a small set of anchors stored exactly, expresses every other token through its most similar anchor, and refines only those whose approximation most affects the model's output. AnchorKV consistently preserves accuracy across models and datasets, retaining 99\% of the full-cache score at the 70B scale, while keeping the entire context at a fraction of its cost.
\end{abstract}
\section{Introduction}
\begin{figure*}[t]
\centering
\includegraphics[width=0.85\textwidth]{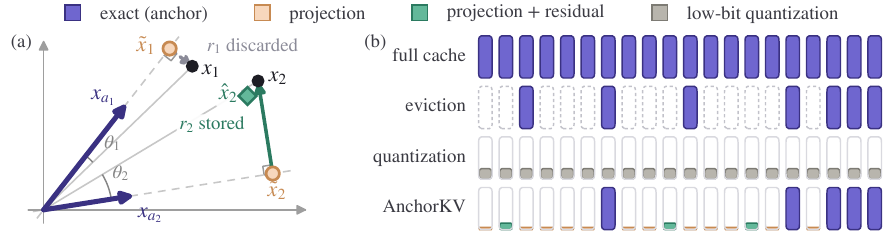}
\caption{Overview of AnchorKV. (a) The representation in $\mathbb{R}^2$: anchors are stored exactly as-is, and every other token is projected onto the span of its nearest anchor and encoded by an anchor index and one coefficient $\gamma_i = \langle x_i, x_{a(i)} \rangle / \|x_{a(i)}\|^2$, which absorbs the anchor's norm. Token $x_1$ aligns closely with its anchor and stores no residual; $x_2$ aligns poorly, receives a budgeted slot, and stores a 2-bit residual, giving $\hat{x}_2 = \tilde{x}_2 + \tilde{r}_2$. (b) Bytes per token under three compressed regimes at a common budget, against the full cache. Eviction spends everything on exactly-stored tokens and so keeps more of them, while AnchorKV affords fewer anchors because the rest of its budget represents every remaining token.}
\label{fig:overview}
\end{figure*}
Long-context inference is now the common case: models read whole repositories, documents, and conversation histories, and context windows run to hundreds of thousands of tokens. The limit is memory, not compute. The key-value (KV) cache stores one key and one value per token per layer, so it grows linearly with context and batch size and is read in full at every decoding step. At 128K, a single Llama-3.1-8B \citep{llama3} request holds a 16\,GiB cache in bf16, as much as the model weights serving it. The cache, not the model, decides how many requests a device holds and how fast each advances.

Several lines of work attack this problem. The dominant one is eviction, which scores tokens and keeps only a subset, dropping the rest \citep{h2o, snapkv, pyramidkv, adakv}. A dropped token costs nothing, so compression is unbounded in principle, and the highest compressions come from these works. The difficulty is that scoring happens before the queries that will read the cache are known: a token the prompt's last queries ignored is gone when a later query needs it, and our per-task results show this is not rare but the family's dominant failure mode (Section~\ref{subsec:main-results}). Offloading and retrieval systems answer this by holding the evicted state in host memory and fetching the blocks a query needs \citep{infllm, quest}. Reach is restored, but the full cache still lives somewhere, and the fetch and its bookkeeping now sit in the decode loop, trading a memory problem for a transfer and scheduling one.

A second line quantizes, keeping every token resident in fewer bits \citep{kivi2024, kvquant, turboquant}. Nothing is ever out of reach, the property eviction gives up. What this line gives up instead is depth: accuracy falls as the representation is pushed to one or two bits, so compression is bounded. In our comparison, it stops near $5\times$, short of eviction's ratios and of the regime where the cache stops being the binding constraint.

In this work we present AnchorKV, which keeps every token and still compresses aggressively, storing the cache in an anchor-residual representation. Each KV head stores a small set of anchor tokens exactly and represents every other token as a scaled copy of its closest anchor, at a cost of a few bytes instead of a full vector (Figure~\ref{fig:overview}). Whatever the byte budget leaves buys quantized residuals for the tokens whose approximation costs the most attention-output error. No position leaves the softmax or is out of reach of a later query, so the full context survives at a fraction of its cost and the compressed cache is a perturbation of the exact one, not a truncation of it.

In AnchorKV, compression runs once at the end of prefill, on a frozen model, leaving the attention arithmetic unchanged. The user sets a single knob: the fraction of the cache to retain, and an exact byte accounting turns it into a storage plan that matches the requested budget; every other parameter is held fixed across all models and benchmarks. Under that one setting, AnchorKV at $20\times$ scores higher than every eviction baseline at $10\times$, the same accuracy at half the memory, across three model scales and three long-context benchmarks. On Llama-3.1-70B it retains 99.3\% of the uncompressed RULER \citep{ruler} score at $20\times$ against 86.8\% for the strongest baseline, and retention improves with scale, the regime where the cache is most expensive to hold.

Section~\ref{sec:related} reviews prior compression methods, Section~\ref{sec:method} presents AnchorKV, and Section~\ref{sec:experiments} evaluates it at matched byte budgets against eviction and quantization baselines, ablates each component, and reports runtime and memory.

\section{Related Work}
\label{sec:related}

Long-context inference has made the KV cache a standard target for compression, and the work on it falls into a few recognizable families. We review three: token eviction, quantization and low-rank projection, and shared representations.

\paragraph{Token Eviction.} The dominant approach keeps a subset of tokens and drops the rest, using attention sinks with a recency window \citep{streamingllm}, accumulated attention mass \citep{h2o, scissorhands, tova}, or an observation window that scores the prompt at prefill, such as SnapKV \citep{snapkv}. Later work keeps the rule and improves the budget, distributing capacity across layers \citep{pyramidkv} or heads \citep{adakv}, or ranking tokens by their change in attention output rather than by attention score alone \citep{criticalkv, obcache}. AnchorKV uses the same prefill-time scoring machinery, the SnapKV observation window in particular, but spends it on a different decision: how many bytes a token receives, not whether it is kept. The binary decision of eviction is irreversible, and the cost is visible in our per-task results, where the eviction baselines collapse on exactly the tasks whose answer tokens the observation window did not score highly (Section~\ref{subsec:main-results}).

\paragraph{Quantization and Low-Rank Projection.} An orthogonal line keeps every token and compresses its representation, either lowering the bit width of each entry through per-channel and outlier-aware schemes \citep{kivi2024, kvquant, kang2024gear, kvlinc2025}, or projecting keys and values onto a low-rank subspace across layers \citep{minicache, xkv} or along the hidden dimension \citep{palu}. AnchorKV shares the commitment to the full context, and departs on how far compression can be pushed: accuracy in this line falls off as the representation approaches one or two bits, which caps it near $5\times$ in our comparison. AnchorKV instead varies how many bytes a token receives, and reaches $20\times$. The two axes are complementary: our residual codec already uses a randomized rotation of the kind this line introduced.

\paragraph{Shared Representations.}
A third family represents several tokens together, merging evicted states into retained neighbors \citep{zhang2024cam, wan2024d2o}, grouping similar tokens \citep{kvmerger}, or coding tokens against a shared set of centroids or codebook entries \citep{commvq, centroidkv2025, liu2024clusterkv, hooper2024squeezed}. AnchorKV is closest to this family, since an anchor is a reference vector and most tokens are coded against one. What differs is what survives the coding: when tokens collapse onto a shared representation, the softmax can no longer separate them. AnchorKV gives every token its own coefficient along its anchor, so no two become identical, and adds a residual for those whose approximation costs the most output error.

\section{Method}
\label{sec:method}
AnchorKV compresses the KV cache without removing positions from attention. At the end of prefill, each KV head selects $k$ \textit{anchor} positions and stores their keys and values exactly. Every other vector is approximated by its projection onto the span of a single anchor (Figure~\ref{fig:overview}), costing only an anchor index and a scalar coefficient. Tokens whose projection costs the most attention-output error also store a quantized \textit{residual}. The user fixes $\theta$, the fraction of the uncompressed cache to retain, and an exact byte accounting turns it into the number of residuals a layer can store.

Throughout, $S$ is the context length, $D$ the per-head dimension, $H$ and $H_q \geq H$ the number of KV and query heads, $W$ the recency window, and $k$ the per-head anchor budget, which includes the window. All operations are per layer and per KV head, whose keys and values we write as $K, V \in \mathbb{R}^{S \times D}$.

\subsection{Anchor-Residual Representation}
\label{subsec:arc}
The representation applies independently to $K$ and to $V$, so we describe it for generic vectors $x_1, \dots, x_S \in \mathbb{R}^D$ standing for the rows of either. A subset $\mathcal{A}$ of them, the \textit{anchors}, is stored exactly, and each anchor provides a direction along which every other vector is represented. Each non-anchor vector is assigned to the anchor whose direction is closest to its own,
\begin{equation}
a(i) = \argmax_{a \in \mathcal{A}} \frac{\left|\langle x_i, x_a \rangle\right|}{\lVert x_i \rVert \lVert x_a \rVert},
\label{eq:nearest-anchor}
\end{equation}
and approximated by its projection onto that anchor,
\begin{equation}
\gamma_i = \frac{\langle x_i, x_{a(i)} \rangle}{\lVert x_{a(i)} \rVert^2},
\qquad
\tilde{x}_i = \gamma_i x_{a(i)},
\qquad
r_i = x_i - \tilde{x}_i,
\label{eq:projection}
\end{equation}
where $\tilde{x}_i$ is the orthogonal projection of $x_i$ onto the span of $x_{a(i)}$ and $r_i$ is the residual the projection discards.

Thus, every non-anchor vector is represented by an anchor index $a(i)$ and a
scalar coefficient $\gamma_i$. For a selected subset of positions
$\mathcal{R}$, AnchorKV additionally stores a quantized residual
$\tilde{r}_i$. The reconstructed vector is
\begin{equation}
\hat{x}_i
=
\tilde{x}_i
+
\mathbf{1}[i \in \mathcal{R}]\,\tilde{r}_i.
\label{eq:reconstruction}
\end{equation}
Without a stored residual, the vector is reconstructed only from its anchor
projection. Storing the residual restores the token-specific component that
lies outside the anchor direction.

We apply this representation independently to keys and values. They share the
same anchor positions, but use separate assignments, projection coefficients,
and residual sets $\mathcal{R}_K$ and $\mathcal{R}_V$.

% AnchorKV replaces the dense per-head caches $K, V \in \R^{S \times D}$, applying the same representation independently to keys and to values; we write it for a generic set of vectors $x_1, \dots, x_S$. An anchor selector returns a subset $\mathcal{A}$ of size $k$, stored exactly (Section~\ref{subsec:anchorkv}). Every other vector is stored as the index of the anchor whose line passes closest to it, equivalently the anchor of highest absolute cosine similarity,
% \begin{equation}
% a(i) = \argmax_{a \in \mathcal{A}} \frac{\abs{\inner{x_i}{x_a}}}{\norm{x_a}},
% \label{eq:nearest-anchor}
% \end{equation}
% together with the multiple of that anchor that comes closest to it,
% \begin{equation}
% \gamma_i = \frac{\inner{x_i}{x_{a(i)}}}{\norm{x_{a(i)}}^2},
% \quad
% \tilde x_i = \gamma_i\, x_{a(i)},
% \quad
% r_i = x_i - \tilde x_i,
% \label{eq:projection}
% \end{equation}
% so $\tilde x_i$ is the orthogonal projection of $x_i$ onto its anchor's line and $r_i$ is the \textit{residual} the projection discards (Figure~\ref{fig:overview}). Residuals are stored, quantized, for a subset $\mathcal{R}$ of the non-anchors, giving
% \begin{equation}
% \hat x_i \;=\; \tilde x_i \;+\; \1[i \in \mathcal{R}]\; \tilde r_i .
% \label{eq:reconstruction}
% \end{equation}
% Keys and values carry their own sets $\mathcal{R}_K$ and $\mathcal{R}_V$, sized in Section~\ref{subsec:allocation} and filled by how much output error their residuals recover (Section~\ref{subsec:utility}).

\paragraph{Residual quantization.}
Stored residuals are quantized at two bits per coordinate. Before quantization, a randomized Hadamard transform spreads the residual energy across coordinates, reducing sensitivity to outliers \citep{ailon2009fast,quip2024,quarot2024}. Each rotated residual is normalized using a per-token absmax scale and quantized with the four-level Lloyd--Max
codebook for a unit Gaussian source \citep{lsq1957,quantizing1960}. The codes are packed four per byte, so each stored residual requires $D/4$ bytes for its coordinates plus one scale. Appendix~\ref{app:codec} specifies the codec and storage layout.

\subsection{AnchorKV Cache Compression}
\label{subsec:anchorkv}
We now describe how AnchorKV uses this representation to compress the cache. Compression runs once, at the end of prefill, in three steps. Each KV head first selects its $k$ anchor positions. Every other position is then assigned to an anchor and projected onto it, separately on the key and the value side. Finally, a subset of positions receives a quantized residual, allocated where it recovers the most output error. Heads select and assign on their own anchors, but compete for a single residual budget per layer (Section~\ref{subsec:allocation}). Anchors fix both what is stored exactly and the directions everything else is represented along, so we begin there.

\paragraph{Anchor selection.}
The final $W$ positions are always stored exactly and count against the anchor budget. They serve three purposes at once: the recency window, the observation queries that score earlier positions, and anchor directions that cost no extra slots. Each earlier position is scored as in SnapKV \citep{snapkv}, by the attention it receives from those queries averaged over the query heads of its KV head, pooled over a positional kernel of width $\kappa$ so a sharp peak also promotes its neighbors (Appendix~\ref{app:selector}).
A fraction $\rho$ of the remaining $k - W$ slots goes to the highest scores, and the rest is sampled uniformly from the context. Attention-based selection identifies positions that the observation queries use directly. However, an anchor also serves as a direction onto which other vectors are projected, and high-attention tokens do not necessarily provide good directional coverage of the cache. The uniformly sampled anchors improve this coverage.

% \paragraph{RoPE.}
Key projection is performed before RoPE because positional rotation weakens the directional similarity used for anchor assignment. Anchor scoring still uses post-RoPE keys, so it reflects the attention computed by the model.
RoPE \citep{rope2021} is a per-position linear map $R_t$, so it distributes over the decomposition, $R_t K_t = R_t \tilde{K}_t + R_t r^K_t$: thus keys can be stored before rotation and rotated during decoding, and values are unaffected.

\subsection{Attention-Output-Aware Residual Scoring}
\label{subsec:utility}
Neither the size of a residual nor the attention a position receives determines how much storing it helps: a large residual on a barely attended position changes the output little, and so does heavy attention on a position its anchor already represents well. AnchorKV therefore ranks residuals by their first-order effect on the attention output.

Given a layer, a KV head and a query $q$,
\begin{equation}
s_t = \frac{q^\top K_t}{\sqrt{D}},
\qquad
\alpha = \operatorname{softmax}(s),
\qquad
y = \sum_t \alpha_t V_t
\label{eq:exact-attention}
\end{equation}
denote the exact logits, attention weights, and output. Write $\hat{\alpha}$ and $\hat{y}$ for the same quantities computed from the reconstructed cache, and set
$\Delta K_t=\hat{K}_t-K_t$,
$\Delta V_t=\hat{V}_t-V_t$, and
$\Delta s_t=q^\top\Delta K_t/\sqrt{D}$. 
For a token without a stored residual, $\Delta K_t=-R_t r_t^K$ and $\Delta V_t=-r_t^V$; otherwise, only quantization error remains.
Let $\overline{\Delta s}=\sum_t\alpha_t\Delta s_t$. Applying the softmax Jacobian gives
\begin{equation}
\Delta y
\approx
\underbrace{
\sum_t
\alpha_t
\left(\Delta s_t-\overline{\Delta s}\right)
\left(V_t-y\right)
}_{\text{K channel}}
+
\underbrace{
\sum_t \alpha_t \Delta V_t
}_{\text{V channel}}.
\label{eq:channels}
\end{equation}
See Appendix~\ref{app:utility} for details. Key errors act through the attention weights and are modulated by $V_t-y$, whereas value errors act directly on the output. We therefore score key and value residuals separately.

Under an incoherence approximation in which per-token contributions are zero-mean and independent, the expected squared norm of Eq.~\ref{eq:channels} decomposes into additive token-wise terms. Since residuals on each side have equal cost, selecting the largest terms minimizes this approximation for a fixed budget. A position's utility is the term it contributes when its residual is dropped, estimated with the $m = W(H_q/H)$ observation queries of its KV head:
\begin{equation}
\begin{aligned}
u_t^K
&=
\frac{1}{m}
\sum_{w=1}^{m}
\left(\alpha_t^{(w)}\right)^2
\frac{
\left(q^{(w)\top}R_t r_t^K\right)^2
}{D}
\left\lVert V_t-y^{(w)}\right\rVert^2,
\\[3pt]
u_t^V
&=
\frac{1}{m}
\sum_{w=1}^{m}
\left(\alpha_t^{(w)}\right)^2
\left\lVert r_t^V\right\rVert^2.
\end{aligned}
\label{eq:utilities}
\end{equation}
Here, $q^{(w)}$, $\alpha^{(w)}$, and $y^{(w)}$ are the query, exact attention distribution, and exact output for observation query $w$.
Utilities are comparable across KV heads within the same side, enabling the pooled allocation in Section~\ref{subsec:allocation}. All required quantities are available at the end of prefill, so utility estimation requires a single pass over the cache.

Like observation-window eviction methods \citep{snapkv,pyramidkv,adakv}, the estimator assumes that the positions recent queries use are the positions decoding will need. What differs is the consequence of a mistaken estimate: a position AnchorKV prices low is stored coarsely and remains available to later queries.

\paragraph{Why retain every token?}
The utility scores determine where residuals are spent, but the anchor-residual representation provides a separate guarantee that holds regardless of the allocation. Since no position leaves the softmax, the compressed cache is a perturbation of the exact one, and adding and subtracting $\sum_t \hat{\alpha}_t V_t$ splits the output error into
\begin{equation}
\lVert y-\hat{y}\rVert \leq E_K + E_V,
\label{eq:output-bound}
\end{equation}
with $E_K = \lVert\sum_t(\alpha_t-\hat{\alpha}_t)V_t\rVert$ induced by the reconstructed keys and $E_V = \sum_t\hat{\alpha}_t\lVert V_t-\hat{V}_t\rVert$ the attention-weighted value error. Anchors are exact, so the perturbation is driven entirely by non-anchor positions. Appendix~\ref{app:proof1} bounds $E_K$ by the largest key-induced logit perturbation, so key error is controlled by a worst case while value error is averaged under attention mass.

On the key side, retention is not automatically an advantage: a poorly reconstructed key can attract attention mass that its exact counterpart would not receive, which eviction avoids by removing the position outright. A fixed number of residuals cannot control a maximum over positions, so this term is controlled by the representation instead. Every key keeps its own projection, so no key error exceeds that position's own residual norm.

The value side is where the comparison with eviction is sharp. Write $\lVert V_t-\hat{V}_t\rVert = \beta_V(t)\lVert r_t^V\rVert$, with $\beta_V(t) = 1$ when no residual is stored and, by Jensen's inequality, $\E[\beta_V(t)] \leq \sqrt{\eta}$ when one is, where $\eta$ is the normalized mean-squared distortion of the quantizer. Then
\begin{equation}
E_V
\leq
\left(
    \max_{t \notin \mathcal{A}}
    \beta_V(t)
    \frac{
        \lVert r_t^V\rVert
    }{
        \lVert V_t\rVert
    }
\right)
\sum_{t \notin \mathcal{A}}
\hat{\alpha}_t
\lVert V_t\rVert.
\label{eq:mass-alignment}
\end{equation}
The sum on the right is the value mass outside the anchor set, the term an eviction method with the same kept set pays for discarding those positions. AnchorKV pays it scaled by the largest relative projection error left unrepaired, a factor at most one, and one that shrinks as non-anchor positions gain nearby anchors or receive residuals. The uniform share of anchor selection buys exactly this coverage. The comparison is not exact, since the weights in Eq.~\ref{eq:mass-alignment} are the reconstructed $\hat{\alpha}$ and eviction additionally renormalizes the softmax over a smaller support.

\subsection{Byte-Budgeted Residual Allocation}
\label{subsec:allocation}
The retained fraction $\theta$ fixes the byte budget of every layer. Anchors and per-token metadata are committed first, and whatever remains buys residuals:
\begin{equation}
N
=
\max\left\{
0,\,
\left\lfloor
\frac{\theta M_\mathrm{full} - M_{\mathrm{base}}}{b_{\mathrm{res}}}
\right\rfloor
\right\},
\label{eq:byte-budget}
\end{equation}
where $M_\mathrm{full}$ is the uncompressed cache of the layer, $M_{\mathrm{base}}$ the bytes held by its anchors and metadata, and $b_{\mathrm{res}}$ the bytes of one quantized residual. Appendix~\ref{app:codec} gives the complete accounting, which charges every stored tensor, so the compressed footprint never exceeds the budget.
Keys receive $\floor{N/2}$ slots and values the rest.
Within each side, we select the highest-utility residuals according to Eq.~\ref{eq:utilities}, pooling candidates across all KV heads in the layer.
This allows heads with larger estimated output errors to receive more of the residual budget.
The retained fraction $\theta$ is the only user-facing compression parameter. We set the anchor budget $k$ as a fixed fraction of the context length and keep $W$, $\rho$, and $\kappa$ fixed across layers and experiments; their values are reported in Section~\ref{sec:experiments}.

\paragraph{Storage and Decoding}
\label{subsec:storage}
Anchors are stored in bf16, keys before RoPE. Every other position stores an anchor index and a coefficient per side, and positions in $\mathcal{R}_K$ or $\mathcal{R}_V$ add a packed residual and its scale, laid out sparsely so positions without one cost nothing. The dense $K, V$ are discarded and never rebuilt: keys are reconstructed by Eq.~\ref{eq:reconstruction} and rotated by $R_i$, values by Eq.~\ref{eq:reconstruction} alone, and generated tokens are appended exactly. Reconstruction is fused into a FlashAttention-style tiled kernel \citep{dao2024flashattention2}, so each tile reconstructs only the keys and values it consumes and the dense cache is never materialized: the runtime footprint is the compressed footprint. Section~\ref{subsec:efficiency} reports memory measurements and decoding throughput.

% \paragraph{Parameters.}
% The target ratio $\theta$ is the only user-facing knob. The anchor budget $k$ is a fixed fraction of the context length, includes the window $W$, and is constant across layers. It, together with $W$, the scored fraction $\rho$ and the pooling kernel $\kappa$, is held fixed across all experiments, with defaults and ablations in Section~\ref{sec:experiments}.
\section{Experiments}
\label{sec:experiments}
We evaluate AnchorKV across three model scales and three long-context benchmarks against eviction and quantization baselines at matched byte budgets (Sections~\ref{subsec:exp-setup} and \ref{subsec:main-results}), ablate each design choice and analyze the internal behavior of the compressed cache (Section~\ref{subsec:ablations}), and measure the throughput and memory cost of the method (Section~\ref{subsec:efficiency}).

\begin{figure*}[t]
\centering
\includegraphics[width=0.85\textwidth]{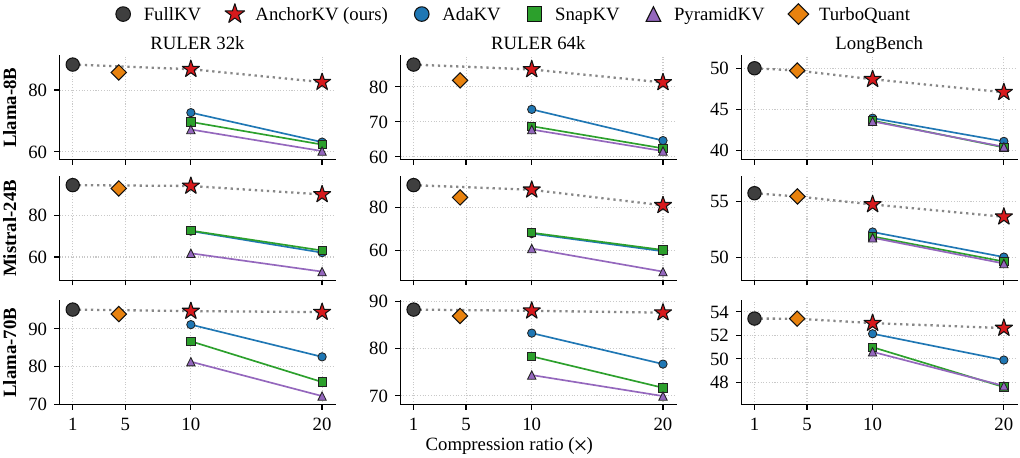}
\caption{AnchorKV against eviction and quantization baselines at matched byte budgets. Rows are Llama-3.1-8B, Mistral-Small-3.1-24B and Llama-3.1-70B; columns are RULER-32K, RULER-64K and LongBench. At both $10\times$ and $20\times$, AnchorKV is the highest-scoring compressed cache in every panel, and its score at $20\times$ exceeds that of every baseline at $10\times$, so the same accuracy is reached with half the memory. TurboQuant is measured at 3.5 bits and stays close to the uncompressed cache, but its compression is bounded below $5\times$.}
\label{fig:main-results}
\end{figure*}

\subsection{Setup}
\label{subsec:exp-setup}
\paragraph{Models and benchmarks.}
We evaluate three instruction-tuned models spanning an order of magnitude in scale: Llama-3.1-8B-Instruct and Llama-3.1-70B-Instruct \citep{llama3}, and Mistral-Small-3.1-24B-Instruct \citep{mistralsmall31}, all from public checkpoints with their default chat templates.
RULER \citep{ruler} supplies 13 synthetic tasks that evaluate long-context capabilities, which we run at 32K and 64K. LongBench \citep{longbench} covers various long-context tasks with real documents (e.g., summarization), leaving less redundancy to exploit. As a stress test we sweep context length and needle depth on Needle-in-a-Haystack retrieval \citep{niah}, using a 64-digit passkey since shorter ones are close to saturated for every method we compare against.

\paragraph{Baselines.}
We compare AnchorKV against the uncompressed bf16 cache (FullKV), against three eviction methods, SnapKV \citep{snapkv}, PyramidKV \citep{pyramidkv} and AdaKV \citep{adakv}, which allocate a fixed, layer-adaptive and head-adaptive budget respectively, and against TurboQuant \citep{turboquant}, a rotation-based quantizer that, like AnchorKV, keeps every token. The eviction baselines run through KVPress \citep{kvpress} with its default settings. The two families expose incomparable knobs, a token budget and a bit width, so we convert both to bytes and match all methods at the same per-layer footprint. An eviction method keeping $B$ tokens stores $4BHD$ bytes, since kept tokens remain in bf16 and discarded ones cost nothing. For AnchorKV the footprint is the accounting of Eq.~\ref{eq:byte-budget}, which charges every stored tensor including all per-token metadata. All ratios are relative to the bf16 cache and run up to $20\times$.

\paragraph{Implementation.}
AnchorKV uses one configuration for all models and all tasks: window $W = 32$, anchor budget $k = S/128$, scored fraction $\rho = 0.7$, pooling kernel $\kappa = 7$, anchors in bf16 and residuals at 2 bits. Compression runs once at the end of prefill and decoding is greedy. Experiments use NVIDIA A100-80GB GPUs at batch size 1. Baseline configurations, and benchmark details are in Appendix~\ref{app:setup}.
\begin{figure}[!tb]
\centering
\includegraphics[width=0.9\columnwidth]{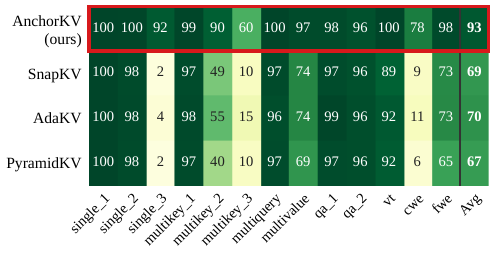}
\caption{Per-task RULER accuracy on Llama-3.1-8B-Instruct at $20\times$, averaged over RULER-32K and 64K and given as a percentage of the uncompressed score. AnchorKV matches or exceeds the best eviction baseline on most tasks and never falls below 60\%, while the baselines drop under 16\% on the tasks that require retrieval against distractors or aggregation over the whole context.}
\label{fig:heatmap-8b}
\end{figure}
\subsection{Main Results}
\label{subsec:main-results}
Figure~\ref{fig:main-results} compares all methods at matched byte budgets across nine model--benchmark settings: three model scales evaluated on RULER-32K, RULER-64K, and LongBench. Two observations hold in every panel. First, AnchorKV stays near the uncompressed cache as compression increases, while the eviction baselines degrade sharply. Second, AnchorKV at $20\times$ outperforms every eviction baseline at $10\times$: the same accuracy is reached with half the cache memory. TurboQuant also tracks FullKV closely, but its fixed 3.5-bit representation caps it near $5\times$, well short of the regime AnchorKV operates in.
The cost of compression also falls with model scale. On RULER-32K at $20\times$, AnchorKV retains $93.5\%$, $95.3\%$, and $99.3\%$ of the FullKV score on Llama-8B, Mistral-24B, and Llama-70B respectively; at 70B the compressed cache scores $94.4$ against $95.0$ uncompressed, while the strongest eviction baseline retains only $86.8\%$. The trend is not simply a property of stronger models: on Mistral-24B the best eviction baseline stays under $67\%$ at both context lengths, no better than at 8B. LongBench shows the same pattern on real documents, with AnchorKV retaining $94.1\%$ at 8B and $98.4\%$ at 70B, ahead of every baseline at every scale. AnchorKV is therefore most accurate precisely where the cache is most expensive to hold. Complete per-dataset results are reported in Appendix~\ref{app:results}.

\paragraph{Per-task consistency.} Figure~\ref{fig:heatmap-8b} breaks RULER down by task at $20\times$. AnchorKV matches or exceeds the strongest eviction baseline on twelve of thirteen tasks and trails it by a single point on the thirteenth. Its floor across the suite is $60\%$ of FullKV, while the strongest baseline drops below $70\%$ on four tasks and below $16\%$ on three. The largest margins appear exactly where eviction's irreversible decision is most costly, retrieval against distractors and aggregation over the whole context: AnchorKV retains $78\%$, $92\%$, and $60\%$ on \texttt{cwe}, \texttt{single\_3}, and \texttt{multikey\_3}, exceeding the strongest eviction baseline by $67$, $88$, and $45$ percentage points, respectively. Additional task-level results are given in Appendix~\ref{app:results}.

\paragraph{Needle-in-a-haystack.}
\begin{figure}[!tb]
\centering
\includegraphics[width=0.95\columnwidth]{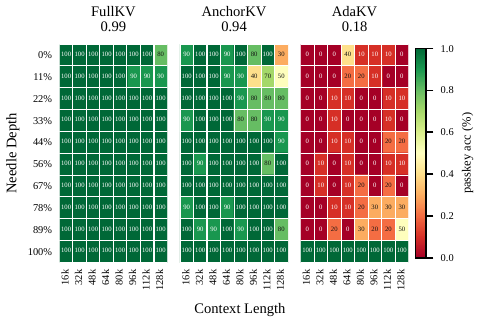}
\caption{Needle-in-a-haystack retrieval with a 64-digit passkey on Llama-3.1-8B-Instruct at $5\times$, over context lengths from 16K to 128K and needle depths from 0\% to 100\%, with mean accuracy above each panel. AnchorKV recovers the needle across nearly the whole grid at 0.94 against 0.99 for the uncompressed cache, while the best eviction baseline (AdaKV) averages 0.18. All methods are exact along the bottom row, where the needle falls inside the recency window they store verbatim.}
\label{fig:niah}
\end{figure}
 Figure~\ref{fig:niah} sweeps context length and needle depth at $5\times$. AnchorKV averages 0.94 against 0.99 for the uncompressed cache and recovers the needle across nearly the whole grid, while the strongest eviction baseline, AdaKV, averages 0.18 and fails almost everywhere: a needle the observation window did not score is simply gone. AnchorKV's few soft cells sit at shallow depths in the longest contexts, where a needle placed early is furthest from the queries that score the anchors. Full grids at $5\times$ and $10\times$ for all baselines are in Appendix~\ref{app:results}.
\begin{figure}[!tb]
  \centering
  \includegraphics[width=\columnwidth]{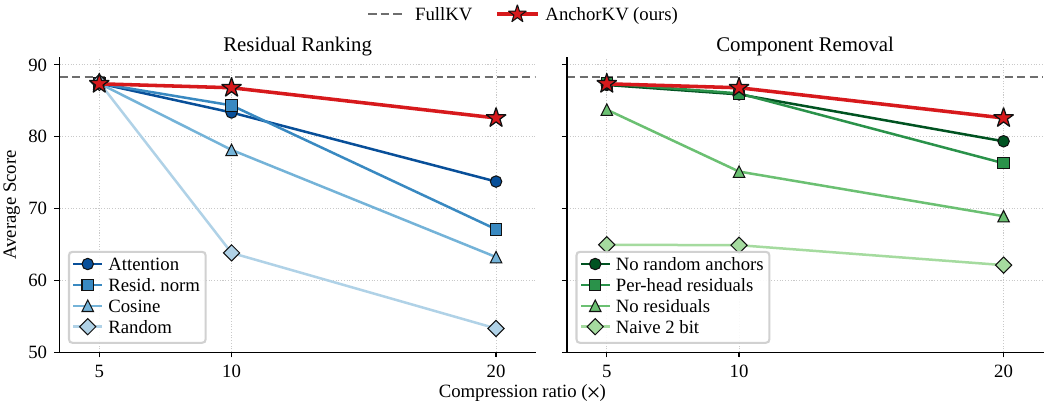}
  \caption{Ablation Study. Evaluated on Llama-3.1-8B, RULER-32K, at ×5/×10/×20 (one variable changed at a time, byte budget fixed). Left: how residuals are ranked; Right: component removals. AnchorKV leads every variant, and the gaps widen as the budget tightens.}
  \label{fig:ablations}
\end{figure}

% \begin{figure*}[t]
% \centering
% \includegraphics[width=\textwidth]{Images/performance.pdf}
% \caption{Systems behaviour of AnchorKV at a $20\times$ target
% (Llama-3.1-8B-Instruct, one 80\,GiB A100, bf16).
% \textbf{(a)} Resident KV cache after prefill at $B{=}1$, excluding weights;
% the reduction is $18.5\times$ at 16K and $19.5\times$ at 128K, approaching the
% target from below as the fixed per-token metadata is amortised.
% \textbf{(b)} Decode latency at $B{=}1$. The fused kernel pays a fixed
% reconstruction cost per step, so AnchorKV is slower at short context, but the
% cost amortises as the cache shrinks relative to the context and the curves cross
% near 96K.
% \textbf{(c)} Aggregate throughput at 64K against batch size. The full cache is
% out of memory at $B{=}4$, where AnchorKV still gains.}
% \label{fig:systems}
% \end{figure*}
\begin{figure*}[t]
\centering
\includegraphics[width=0.9\textwidth]{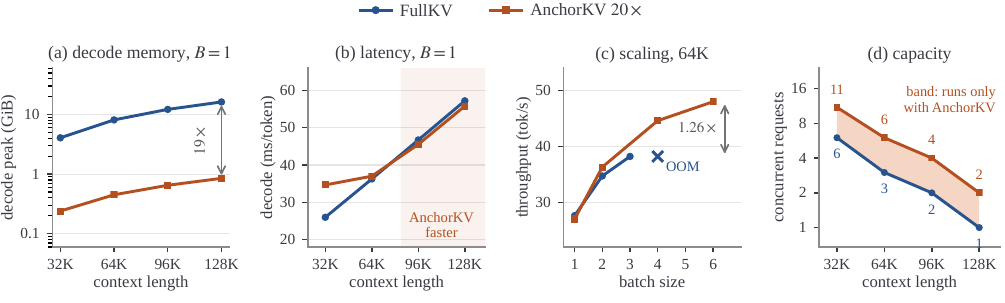}
\caption{Efficiency Analysis. We profile AnchorKV at a $20\times$ target
(Llama-3.1-8B-Instruct, bf16, one 80\,GiB A100; baseline is the full cache under PyTorch SDPA, memory excluding weights).
(a) Decode peak memory per request, log scale: $17$--$19\times$
smaller.
(b) Decode latency. The fused kernel costs $1.3\times$ at 32K, reaches
parity near 64K, and is $2$--$3\%$ faster at 96K--128K (shaded).
(c) Throughput against batch size at 64K, each arm walked to its memory
ceiling; $\times$ marks the first batch that fails. AnchorKV sustains $B{=}6$
against $B{=}3$, a $1.26\times$ gain between frontiers.
(d) Largest batch that fits. The band is the set of configurations only
AnchorKV can serve: $2\times$ the concurrency from 64K upward.}
\label{fig:systems}
\end{figure*}

\subsection{Ablations and Analysis}
\label{subsec:ablations}

We now isolate the two decisions that define AnchorKV: how anchors are chosen and how residuals are spent. All ablations use Llama-3.1-8B on RULER-32K, change one choice at a time, and hold the byte budget fixed. Each variant runs at $5\times$, $10\times$ and $20\times$, so Figure~\ref{fig:ablations} shows both the effect of a choice and how that effect grows as the budget tightens.

\paragraph{Residual Ranking.} We measure the effect of the metric that decides which tokens receive a residual, while leaving the anchors and the budget untouched. In place of the utility of Eq.~\ref{eq:utilities} we rank by (i) cosine similarity to the nearest anchor, a measure of alignment, (ii) residual norm, a measure of approximation error, (iii) attention score, a measure of how heavily a token is used, and (iv) at random, a check that targeting is needed at all. The utility leads at every ratio and random ranking is the worst everywhere. At $5\times$ the choice barely matters, since the budget reaches almost every token and there is little left to choose between. The gaps open as the budget tightens and are widest at $20\times$, where random placement falls far behind the rest. What separates the alternatives there is how much of the utility each one recovers. Each alternative prices one factor and ignores the other. Attention sees how heavily a token is used but not how badly its anchor represents it, the residual norm sees the reverse, and cosine sees neither, only the angle between the two.
The utility is a first-order estimate of the attention output error, and Appendix~\ref{app:attn_out} measures that error directly, so the quantity the ranking is designed to control can be read off on its own rather than only through downstream accuracy.

\paragraph{Component Removal.} We remove one component at a time and measure the change in average score. We (i) select all anchors by attention score, dropping the random share, (ii) give every head an equal residual budget, dropping the pooling across heads, (iii) store no residuals, spending the whole budget on anchors, which leaves projection onto the dictionary alone, and (iv) quantize residuals at 2 bits without the Hadamard rotation and the Lloyd--Max codebook. Every removal costs accuracy. Naive quantization is the largest loss and stays roughly constant across ratios, since it degrades every stored residual rather than changing how many are stored. Storing no residuals costs accuracy at all three ratios even though the freed bytes buy more anchors: a larger dictionary does not replace what a residual restores. Equal per-head budgets and attention-only anchors cost little at $5\times$ and more as the budget tightens, the same pattern the ranking metrics show. Appendix~\ref{app:head_residuals} maps the share of residuals each head receives: the shares are far from uniform and shift with the input, so an equal split cannot follow them. Both halves of the figure point the same way: slack hides these choices, and each matters in the regime the method targets, where every byte must go where it buys the most accuracy.

\begin{figure}[!tb]
\centering
\includegraphics[width=0.85\columnwidth]{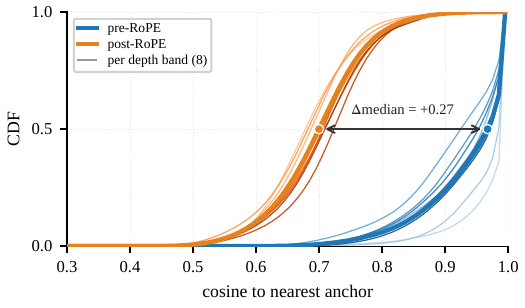}
\caption{Pre-RoPE vs. Post-RoPE Key Projection. We measure the per-token cosine to the nearest anchor on Llama-3.1-8B at RULER 32K, before and after RoPE. The two labeled curves pool four tasks, one per RULER category
(\texttt{multikey\_3}, \texttt{vt}, \texttt{fwe}, \texttt{qa\_2}), with 20 samples each; the faint curves split the same data into eight bands of four
consecutive layers. RoPE lowers the median cosine by $0.27$, and the ordering is the same in every band.}
\label{fig:rope-cdf}
\end{figure}

\textbf{Effect of Pre-RoPE.} Finally, we ask if keys should be projected before or after RoPE. Figure~\ref{fig:rope-cdf} reports the cosine to the nearest anchor in each case: the median is $0.27$ lower once the rotation is applied. This matters because a token with no stored residual survives only as a scaled copy of its anchor, so its error is set by how closely an anchor direction aligns with it. The projection is orthogonal, so a cosine $c$ leaves a residual of relative norm $\sqrt{1 - c^2}$: the shift moves the quantity that governs every unrepaired token, both in the utilities of Eq.~\ref{eq:utilities} and the relative value residual of Eq.~\ref{eq:mass-alignment}. The ordering holds in all eight depth bands and four RULER categories, so it reflects the rotation rather than a particular layer or task. Fitting before the rotation is thus better: every unrepaired token sits closer to its anchor, so the same dictionary covers more of the cache at no extra cost. AnchorKV therefore projects keys pre-RoPE and rotates them at decoding.

\subsection{Efficiency and Memory}
\label{subsec:efficiency}
AnchorKV decodes directly from the compressed representation rather than
reconstructing the dense KV cache. Our efficient Triton implementation follows a FlashAttention-style tiled execution pattern \citep{dao2024flashattention2}: each key tile is reconstructed only when consumed, rotated with RoPE, and
contracted with the query in registers, after which only its attention scores are retained. Values are accumulated directly from their anchor projections and sparse residuals; by linearity of the inverse Hadamard transform, the residual contribution is transformed once after accumulation rather than once per token. Reconstructed keys and values are therefore never written to global memory, yielding an approximately $19\times$ reduction in steady-state decode peak memory at a $20\times$ target (Figure~\ref{fig:systems}a). Appendix~\ref{app:systems} provides the kernel and memory-accounting details. On Llama-3.1-8B-Instruct, decode latency is higher at short contexts and comparable from 64K upward (Figure~\ref{fig:systems}b). The smaller footprint also buys concurrency: at 64K it doubles the maximum batch size and, with each method at its memory frontier, improves throughput by $1.26\times$ (Figure~\ref{fig:systems}c-d). Overall, AnchorKV pairs a much smaller decode footprint with competitive long-context latency and higher throughput.
\section{Conclusion}
We introduced AnchorKV, a KV cache compression method that keeps every token at a fraction of the memory. Each token is stored in an anchor-residual form, and residuals go where they most preserve the attention output. At $20\times$, AnchorKV outscores every eviction baseline at $10\times$, reaching higher accuracy with half the memory. At the 70B scale it retains up to 99.3\% of the uncompressed accuracy. A tiled implementation carries these savings into decoding. Under a fixed memory budget, it supports higher throughput and more concurrent requests. Together, these results show that aggressive compression, full-context reach and practical serving efficiency are not in conflict.


\bibliography{aaai2027}

\newpage
\appendix
\section{Method - Further details}
\label{app:method}

\subsection{Residual Codec and Storage Layout}
\label{app:codec}
This appendix gives the codec summarized in Section~\ref{subsec:arc} and the byte accounting behind Eq.~\ref{eq:byte-budget}.
 
\paragraph{Rotation.}
Residuals are rotated before quantization by
\begin{equation}
U = \tfrac{1}{\sqrt D} \mathcal{H}_D\, \mathrm{diag}(s), \qquad s \in \{\pm 1\}^D,
\label{eq:hadamard}
\end{equation}
with $\mathcal{H}_D$ the Walsh-Hadamard matrix and $s$ a sign pattern drawn once from a fixed seed. $U$ is orthogonal, so it preserves norms and is inverted by its transpose, and it applies in $O(D \log D)$ by the fast Walsh-Hadamard transform. Each coordinate of $U r_i$ is a signed average of the coordinates of $r_i$, so unless the residual energy sits in a few coordinates, each rotated coordinate is approximately Gaussian with variance $\norm{r_i}^2/D$ \citep{ailon2009fast}. The seed is shared by every layer and head, so the transform is not stored.
 
\paragraph{Codebook.}
Each rotated coordinate is divided by the per-token absmax scale and quantized with the four-level Lloyd-Max codebook for a unit Gaussian source \citep{lsq1957, quantizing1960}, which matches the marginals the rotation produces. We write $\mathcal{C}$ for the encoder and
\begin{equation}
\eta = \E[\norm{r_i - \tilde r_i}^2] / \norm{r_i}^2
\label{eq:eta}
\end{equation}
for the relative squared distortion the codec leaves, a property of the codec and not of the token. A non-anchor vector is stored as
\begin{equation}
\big(a(i),\ \gamma_i\big)\ \text{for } i \notin \mathcal{R},
\qquad
\big(a(i),\ \gamma_i,\ \rvq_i\big)\ \text{for } i \in \mathcal{R},
\label{eq:arc-tuple}
\end{equation}
with $\rvq_i$ the packed code, and decoded as $\tilde r_i = U^\top \mathcal{C}^{-1}(\rvq_i)$ before Eq.~\ref{eq:reconstruction} is applied.
 
% \paragraph{Layout.}
% Table~\ref{tab:layout} lists the tensors a compressed layer holds. Anchor positions are shared by the two sides, so a selected token contributes one key vector and one value vector, while the index, coefficient and residual flag are per side. Residuals are held in a sparse layout: a bitmask marks which tokens carry one, and the codes are packed contiguously in position order, so a token without a residual costs one bit and no slot. Summing the data-independent rows gives
% \begin{equation}
% M_{\mathrm{fixed}} = H\Big[4kD + 2(S-k)(b_a + b_\gamma) + 2\lceil S/8 \rceil\Big],
% \label{eq:mfixed}
% \end{equation}
% with $b_a$ the anchor index width and $b_\gamma$ the coefficient width in bytes. The residual rows cost $c = D/4 + 4$ each, which is what Eq.~\ref{eq:byte-budget} divides by.
\paragraph{Layout.}
Table~\ref{tab:layout} lists the tensors a compressed layer holds, charging
every byte the decode kernel touches. Anchor positions are shared by the two
sides, so a selected token contributes one key vector and one value vector.
Because keys are stored before RoPE, original positions are kept explicitly:
an int32 per non-anchor token, shared across heads and sides, and an int64 per
selected anchor. Every non-window token carries an index and a coefficient per
side, anchors included ($\gamma = 1$ on themselves), keeping the per-token
arrays uniform. Residuals are held sparsely: a bitmask of 64-bit words marks
which tokens carry one, the codes are packed in position order, and per-word
prefix counts with per-head offsets make the position$\to$slot lookup $O(1)$.
Writing $P = S-W$ for the non-window positions and summing the data-independent rows gives
\begin{align}
M_{\mathrm{base}} = H\Big[4kD + 8(k{-}W) + 2P(b_a{+}b_\gamma)\Big] \notag \\
 + 24H\big\lceil P/64 \big\rceil + 8(H{+}1) + 4P,
\label{eq:mfixed}
\end{align}
with $b_a$ the anchor index width and $b_\gamma$ the coefficient width in
bytes. A key residual costs $c_K = D/4 + 4$ (codes plus an fp32 scale); a
value residual adds a one-byte chunk-local position, $c_V = D/4 + 5$, since
the value pass visits only stored residuals and must map each slot back to its
position, while the key pass walks all positions and reuses the bitmask as its
index. Eq.~\ref{eq:byte-budget} divides by these costs.
 
% \begin{table}[t]
% \centering
% \begin{tabular}{llc}
% \toprule
% Tensor & Shape & Bytes \\
% \midrule
% Anchor keys & $H \times k \times D$ & $2HkD$ \\
% Anchor values & $H \times k \times D$ & $2HkD$ \\
% Anchor index & $2 \times H \times (S{-}k)$ & $2H(S{-}k)b_a$ \\
% Coefficient & $2 \times H \times (S{-}k)$ & $2H(S{-}k)b_\gamma$ \\
% Residual mask & $2 \times H \times S$ bits & $2H\lceil S/8 \rceil$ \\
% \midrule
% Residual codes & $N \times D$ two-bit & $ND/4$ \\
% Residual scales & $N$ & $4N$ \\
% \bottomrule
% \end{tabular}
% \caption{Per-layer storage. The upper block is data independent and forms $M_{\mathrm{fixed}}$; the lower block is what the byte budget of Eq.~\ref{eq:byte-budget} buys.}
% \label{tab:layout}
% \end{table}

\begin{table}[t]
\centering
\small
\setlength{\tabcolsep}{4pt}
\begin{tabular}{llc}
\toprule
Tensor & Shape & Bytes \\
\midrule
Anchor keys & $H \times k \times D$ & $2HkD$ \\
Anchor values & $H \times k \times D$ & $2HkD$ \\
Anchor positions & $H \times (k{-}W)$, int64 & $8H(k{-}W)$ \\
Anchor index & $2 \times H \times P$ & $2HPb_a$ \\
Coefficient & $2 \times H \times P$ & $2HPb_\gamma$ \\
Residual mask & $2 \times H \times \lceil P/64 \rceil$ words & $16H\lceil P/64 \rceil$ \\
Prefix counts & $2 \times H \times \lceil P/64 \rceil$, int32 & $8H\lceil P/64 \rceil$ \\
Head offsets & $H{+}1$ per side, int32 & $8(H{+}1)$ \\
Position ids & $P$, int32 & $4P$ \\
\midrule
Residual codes & $N \times D$ two-bit & $ND/4$ \\
Residual scales & $N$, fp32 & $4N$ \\
Value slot positions & $N_V$, uint8 & $N_V$ \\
\bottomrule
\end{tabular}
\caption{Per-layer storage, with $P = S{-}W$ the non-window positions. The upper
block is data independent and forms $M_{\mathrm{base}}$; the lower block is
what the byte budget of Eq.~\ref{eq:byte-budget} buys, at $c_K = D/4 + 4$ bytes
per key residual and $c_V = D/4 + 5$ per value residual.}
\label{tab:layout}
\end{table}
 
\paragraph{Worked example.}
At $S = 32$K, $D = 128$, $H = 8$, $W = 32$ and $k = S/128 = 256$, the
uncompressed layer is $M_{\mathrm{full}} = 4SHD = 134.2$\,MB. With
$b_a = b_\gamma = 2$, Eq.~\ref{eq:mfixed} commits $3.39$\,MB, or $2.5\%$ of 
the layer, leaving $3.32$\,MB at $\theta = 0.05$ for $N = 91{,}056$ residuals 
out of the $2HP = 523{,}776$ token-sides that hold an assignment. The
committed part grows with $S$ at a fixed rate per token, so the fraction of
the budget available for residuals is close to constant across the context
lengths we evaluate.

%%%%%%%%%%%%%%%%%%%%%%%%%%%%%%%%%%%%%%%%%%%%%%%%%%%%%
\subsection{Anchor Selection}
\label{app:app-selection}
In this section we provide more details on how we choose anchors following SnapKV's setup \cite{snapkv}
\label{app:selector}
\begin{equation}
\mathrm{score}(t) = \mathrm{ap}_\kappa\!\Big(\tfrac{1}{m}\textstyle\sum_{w} \text{sm}\big(q^{(w)\top} K^\top/\sqrt D\big)_t\Big),
\label{eq:selector-score}
\end{equation}

where $\text{ap}$ denotes average pooling, $\text{sm}$ is the softmax function and $m = W(H_q/H)$ where $H$ and $H_q \geq H$ are the number of KV and query heads and $W$ is the window size.

%%%%%%%%%%%%%%%%%%%%%%%%%%%%%%%%%%%%%%%%%%%%%%%%%%%%%
\subsection{Derivation of the Utility}
\label{app:utility}
This appendix derives Eq.~\ref{eq:channels} and the additive estimate behind the utilities of Eq.~\ref{eq:utilities}. Fix a layer, a KV head and a query $q$, and write $s_t = q^\top K_t/\sqrt D$, $\alpha = \softmax(s)$ and $y = \sum_t \alpha_t V_t$.
 
\paragraph{Two channels.}
Compression perturbs $K_t \to K_t + \Delta K_t$ and $V_t \to V_t + \Delta V_t$, and hence the logits by $\Delta s_t = q^\top \Delta K_t/\sqrt D$. The softmax Jacobian is $\partial \alpha_t/\partial s_u = \alpha_t(\delta_{tu} - \alpha_u)$, so to first order
\begin{equation}
\Delta \alpha_t = \alpha_t\big(\Delta s_t - \overline{\Delta s}\big),
\qquad
\overline{\Delta s} = \textstyle\sum_u \alpha_u \Delta s_u,
\label{eq:dalpha}
\end{equation}
which shows that a uniform shift of every logit changes nothing and only the spread of logit errors matters. Expanding $\hat y = \sum_t (\alpha_t + \Delta\alpha_t)(V_t + \Delta V_t)$ and dropping the second-order term gives $\Delta y \approx \sum_t \Delta\alpha_t V_t + \sum_t \alpha_t \Delta V_t$. Since $\sum_t \Delta\alpha_t = 0$, the values may be centered by $y$ without changing the first sum, which yields Eq.~\ref{eq:channels}. Centering is what makes the key channel read in $V_t - y$: a key error re-weights token $t$ against the output the query has already formed, so it is harmless on a token whose value agrees with that output.
 
\paragraph{Additivity.}
Eq.~\ref{eq:channels} prices the whole cache jointly. To turn it into a per-token quantity we adopt an incoherence model: seen against the query and value geometry, per-token errors are zero-mean and independent across tokens, and key errors are independent of value errors. The assumption is that residual directions of distinct tokens are incoherent in high dimension, which the randomized rotation of Eq.~\ref{eq:hadamard} supports for the stored residuals. Under the model the cross terms between distinct tokens and between the two channels vanish in expectation, and the common mode $\overline{\Delta s}$ concentrates near zero, leaving
\begin{equation}
\E\norm{\Delta y}^2 \approx
\sum_t \alpha_t^2\, \E[\Delta s_t^2]\, \norm{V_t - y}^2
+ \sum_t \alpha_t^2\, \E\norm{\Delta V_t}^2 ,
\label{eq:estimate}
\end{equation}
an equality in expectation under the model rather than a bound. Its value is additivity: each token contributes one term, storing a token's residual scales that term and leaves every other term untouched, and every residual costs the same bytes, so minimizing the total under a fixed number of residuals is taking the largest terms.
 
\paragraph{From the estimate to the utilities.}
A non-anchor token with no stored residual carries the dropped residuals, $\Delta K_t = -R_t r^K_t$ and $\Delta V_t = -r^V_t$, so its two terms are $\alpha_t^2 (q^\top R_t r^K_t)^2 \norm{V_t - y}^2 / D$ and $\alpha_t^2 \norm{r^V_t}^2$. Storing the residual replaces the dropped vector by the quantization error, of relative squared energy $\eta$, so the term becomes $\eta$ times as large and the slot gains $(1-\eta)$ of it. The factor is common to every token and affects no ranking, so it is dropped from Eq.~\ref{eq:utilities}. Anchors and window tokens are exact and contribute nothing, so they never compete for slots. Averaging the two terms over the $m = W \cdot (H_q/H)$ observation-window rows $q^{(w)}$ of the head's query group, in place of an expectation over queries that decoding has not yet issued, gives Eq.~\ref{eq:utilities}.
%%%%%%%%%%%%%%%%%%%%%%%%%%%%%%%%%%%%%%%%%%%%%%%%%%%%%%%%%%%%%%%%%

\subsection{Output Error Bound Theorem}
\label{app:proof1}
This appendix states and proves the worst-case bound referenced in Section~\ref{subsec:utility}. It holds for any reconstruction of the keys and values, so it does not depend on how anchors are selected or how residuals are allocated.

Fix a layer, a KV head and a query $q \in \R^D$. Let $K, V \in \R^{S \times D}$ be the exact keys and values used by attention, with keys taken post-RoPE, and let $\hat K, \hat V$ be their AnchorKV reconstructions. The exact and reconstructed logits, attention weights and outputs are
\[
\begin{aligned}
s_t &= q^\top K_t / \sqrt D, & \alpha &= \softmax(s), & y &= \textstyle\sum_t \alpha_t V_t, \\
\hat s_t &= q^\top \hat K_t / \sqrt D, & \hat\alpha &= \softmax(\hat s), & \hat y &= \textstyle\sum_t \hat\alpha_t \hat V_t.
\end{aligned}
\]
Three constants control the bound: the largest exact value norm $V_{\max} = \max_t \norm{V_t}$, the largest key reconstruction error $\Delta_K = \max_t \norm{K_t - \hat K_t}$, and the logit perturbation it induces,
\[
\mu = \frac{\norm{q}\, \Delta_K}{\sqrt D}.
\]

\begin{theorem}[Output error of AnchorKV]
\label{thm:output-error}
The attention output error satisfies
\begin{equation}
\norm{y - \hat y} \;\leq\;
\underbrace{\sum_t \hat\alpha_t\, \norm{V_t - \hat V_t}}_{E_V}
\;+\;
\underbrace{2\, V_{\max}\, \tanh(\mu)}_{E_K}.
\label{eq:output-error-tight}
\end{equation}
Since $\tanh(\mu) \leq \mu$, this implies the linear bound
\begin{equation}
\norm{y - \hat y} \;\leq\; \sum_t \hat\alpha_t\, \norm{V_t - \hat V_t} \;+\; \frac{2\, V_{\max}\, \norm{q}}{\sqrt D}\, \Delta_K.
\label{eq:output-error-linear}
\end{equation}
\end{theorem}

The two terms differ in structure. $E_V$ is an attention-weighted average, so a value error on a token the query barely attends to contributes little. $E_K$ depends only on the worst key error, since one badly reconstructed key can shift the softmax normalization for every token.

Both terms are supported on non-anchor tokens alone: anchors and window tokens are stored exactly, so their reconstruction errors vanish. A non-anchor token whose residual is dropped has key error exactly $\norm{r^K_t}$ and value error exactly $\norm{r^V_t}$, and one whose residual is stored has error $\sqrt\eta$ times these in expectation. Every key error is therefore at most that token's own residual norm, which is what bounds $\Delta_K$. Since RoPE is orthogonal, $\norm{K_t - \hat K_t}$ is the same whether measured before or after the rotation.

\begin{proof}
Add and subtract $\sum_t \hat\alpha_t V_t$:
\[
y - \hat y = \sum_t (\alpha_t - \hat\alpha_t)\, V_t + \sum_t \hat\alpha_t\, (V_t - \hat V_t).
\]
By the triangle inequality,
\[
\norm{y - \hat y} \leq \norm{\sum_t (\alpha_t - \hat\alpha_t)\, V_t} + \norm{\sum_t \hat\alpha_t\, (V_t - \hat V_t)}.
\]
The second term is bounded by
\[
\norm{\sum_t \hat\alpha_t\, (V_t - \hat V_t)} \leq \sum_t \hat\alpha_t\, \norm{V_t - \hat V_t},
\]
which gives $E_V$. For the first term,
\[
\norm{\sum_t (\alpha_t - \hat\alpha_t)\, V_t} \leq \sum_t \abs{\alpha_t - \hat\alpha_t}\, \norm{V_t} \leq V_{\max}\, \norm{\alpha - \hat\alpha}_1.
\]
It remains to control $\norm{\alpha - \hat\alpha}_1$. Let $\epsilon_t = \hat s_t - s_t$. By Cauchy-Schwarz,
\[
\abs{\epsilon_t} = \frac{\abs{q^\top (\hat K_t - K_t)}}{\sqrt D} \leq \frac{\norm{q}\, \norm{K_t - \hat K_t}}{\sqrt D} \leq \mu.
\]
Since $\hat\alpha_t = \alpha_t\, e^{\epsilon_t} / Z$ with $Z = \sum_u \alpha_u\, e^{\epsilon_u}$, and $\epsilon_t \in [-\mu, \mu]$, we have $Z \in [e^{-\mu}, e^{\mu}]$ and therefore
\[
\frac{\hat\alpha_t}{\alpha_t} = \frac{e^{\epsilon_t}}{Z} \in [e^{-2\mu}, e^{2\mu}].
\]
We use the following bound on the distance between two distributions with bounded likelihood ratio. If $p, \tilde p$ satisfy $p(t) / \tilde p(t) \in [e^{-2\mu}, e^{2\mu}]$ for every $t$, then
\begin{equation}
\norm{p - \tilde p}_1 \leq 2 \tanh(\mu).
\label{eq:lr-tv}
\end{equation}
Indeed, $\varphi(r) = (r - 1) / (r + 1)$ is increasing on $(0, \infty)$ with $\varphi(e^{\pm 2\mu}) = \pm \tanh(\mu)$, so $\abs{r - 1} \leq \tanh(\mu)\,(r + 1)$ for every $r \in [e^{-2\mu}, e^{2\mu}]$. Applying this with $r = p(t) / \tilde p(t)$, multiplying by $\tilde p(t)$ and summing over $t$,

\[
\begin{aligned}
\norm{p - \tilde p}_1 = \sum_t \abs{p(t) - \tilde p(t)}
&\leq \tanh(\mu) \sum_t \big(p(t) + \tilde p(t)\big) \\
&= 2 \tanh(\mu).
\end{aligned}
\]
Applying Eq.~\ref{eq:lr-tv} with $p = \alpha$ and $\tilde p = \hat\alpha$ gives $\norm{\alpha - \hat\alpha}_1 \leq 2 \tanh(\mu)$, hence
\[
\norm{\sum_t (\alpha_t - \hat\alpha_t)\, V_t} \leq 2 V_{\max} \tanh(\mu),
\]
which is $E_K$. Combining the two terms proves Eq.~\ref{eq:output-error-tight}, and the linear bound follows from $\tanh(\mu) \leq \mu$.
\end{proof}

%%%%%%%%%%%%%%%%%%%%%%%%%%%%%%%%%%%%%%%%%%%%%%%%%%%%%
\subsection{The Algorithm}
\label{app:algo}

Algorithm~\ref{alg:anchorkv} provides an overview of the approach per layer

\begin{algorithm}[t]
\caption{AnchorKV prefill compression (one layer)}
\label{alg:anchorkv}
\begin{algorithmic}[1]
\REQUIRE $K, V$ per KV head; window $W$; anchor budget $k$; target ratio $\theta$
\FOR{each KV head $h$}
  \STATE $\mathcal{A}_h \gets$ last $W$ $\cup$ top $\rho(k{-}W)$ by Eq.~\ref{eq:selector-score} $\cup$ random rest
  \STATE per side: assign $a(t), \gamma_t$ (Eqs.~\ref{eq:nearest-anchor}, \ref{eq:projection}); form $r^K_t, r^V_t$
  \STATE per side: score $u^K_t, u^V_t$ for $t \notin \mathcal{A}_h$ (Eq.~\ref{eq:utilities})
\ENDFOR
\STATE $N \gets$ Eq.~\ref{eq:byte-budget}; \quad $N_K, N_V \gets \lfloor N/2 \rfloor$
\STATE per side $\star$: $\mathcal{R}_\star \gets$ top $N_\star$ by $u^\star_t$ pooled over heads; rotate, quantize, store (App.~\ref{app:codec})
\STATE discard dense $K, V$
\end{algorithmic}
\end{algorithm}
\section{Experiments - Setup details}
\label{app:setup}
In this appendix, we provide further details regarding the experiments we reported in Section~\ref{sec:experiments}.
All the experiments were conducted in NVIDIA A100-80GB GPUs.

\subsection{Models}
We evaluate three instruction-tuned models spanning an order of
magnitude in scale, all with a 128K-token context window: Llama-3.1-8B-Instruct and Llama-3.1-70B-Instruct \citep{llama3}, and Mistral-Small-3.1-24B-Instruct \citep{mistralsmall31}. We apply each model's native chat template and do not truncate inputs (full-context evaluation), so sequence lengths match the benchmark specification exactly, and we apply the KV compression to all of the layers in each model. Models are loaded in \texttt{bfloat16} with FlashAttention-2 and run under Hugging Face \texttt{transformers} / PyTorch.
All generation is greedy (temperature $0$, no sampling) with batch size $1$,
and every run uses a fixed seed ($42$).
 
\subsection{Benchmarks}

\paragraph{RULER}
RULER is a synthetic long-context benchmark based on NIAH \cite{ruler},  but contains more challenging tasks. Overall it has 13 tasks spanning four categories: retrieval, multi-hop, aggregation and question answering. We follow the setup of the official repository using the tasks default configuration to generate RULER 16K-96K. Each different context length has 1300 randomly generated samples, 100 samples per subtask. Scoring follows RULER: a normalized substring match against the gold string(s); tasks with multiple gold items (\texttt{niah\_multiquery}, \texttt{niah\_multivalue}, \texttt{vt}, \texttt{cwe}, \texttt{fwe}) are scored as the fraction of gold items recovered, and the remainder as an exact recall indicator.

\paragraph{LongBench}
LongBench contains 16 tasks across six different categories: single document QA, multi document QA, summarization, few shot learning, synthetic reasoning and code completion. We follow the setup of SnapKV \cite{snapkv}, and do not perform input truncation.
For each task in the benchmark a task-specific evaluation is used, Exact match for QA tasks, ROUGE for summarization, etc. Finally, following the benchmark's official script, we do not apply a chat template when evaluating on the subtasks TREC, TriviaQA, SAMSum, LCC and RepoBench-P tasks. We use all of the samples in each subtask.

\paragraph{Needle-in-a-haystack}
We follow the template used in \cite{kivi2024}, of the passkey retrieval prompt of \cite{niah1}, but instead of using a 7-digit-passkey, we use a 64-digit-passkey and Paul Graham Essays as the background filler.\\
\texttt{
There is an important info hidden inside a lot of irrelevant text. Find it and memorize them. I will quiz you about the important information there.\\
<text from Paul Graham Essays>\\
The pass key is <64-digit passkey>. remember it. <64-digit-passkey> is the pass key.\\
<more text from Paul Graham Essays>\\
What is the pass key? The pass key is:
}

\paragraph{Prompt formatting.} To isolate the effect of KV compression, the answer cue and query prefix are placed after the chat template and the answer cue is excluded from compression (\texttt{two\_stage\_answer\_cue}); the same prompt construction is used for every method, including the uncompressed FullKV baseline, so all methods see an identical prompt and differ only in how the past-token KV cache is stored. For instance for NIAH, the "The pass key is:" is excluded from the compression step.

\begin{figure*}[t]
\centering
\includegraphics[width=\textwidth]{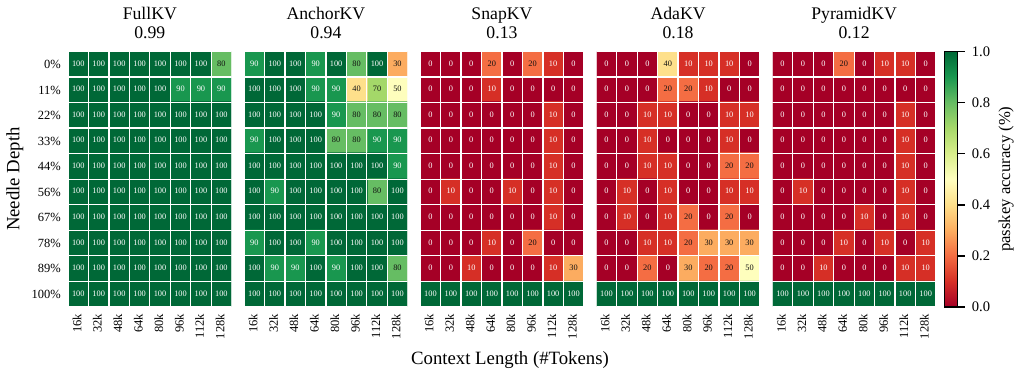}
\caption{Needle in a haystack at $5\times$ compression. Each cell is the mean score over the 10 samples at that context length and needle depth.}
\label{fig:niah-x5-full}
\end{figure*}

\subsection{AnchorKV Hyperparameters}
AnchorKV is applied entirely at inference time; there is no training, fine-tuning, or offline calibration. The method requires no gradient updates and no external calibration corpus: the anchor set, residual selection, and quantization statistics are computed on the fly from the prompt's own key/value tensors during prefill. Consequently there are no learning rates, optimizers, or training schedules to report. The only configuration consists of the fixed inference-time hyperparameters we now list.
For all experiments we use  window $W = 32$, anchor budget $k = S/128$, scored fraction $\rho = 0.7$, pooling kernel $\kappa = 7$, anchors in bf16 and residuals at 2 bits. We swept the anchor budget over $S/64$, $S/128$ and $S/256$ once, on a small subset of samples drawn from a few RULER 
subtasks with Llama-3.1-8B, selected the single best-performing budget, and fixed it for all experiments in the paper; no per-task,
per-benchmark or per-model tuning was performed.

\subsection{Baseline Evaluation Setup}
We compare against three training-free KV-cache eviction baselines, %
\textbf{SnapKV}~\cite{snapkv}, \textbf{AdaKV} (adaptive per-head budget
allocation on top of SnapKV scores)~\cite{adakv}, and
\textbf{PyramidKV}~\cite{pyramidkv}, all evaluated through the official
KVPress framework~\cite{kvpress}. Every baseline is run on the \emph{identical} prompts, compression budgets, decoding settings, and scoring code as our method; the only difference is the eviction rule.
All the default hyperparameters set by KVPress are used. $W=64$ and $\kappa=5$ for the three methods. 
Finally, all the methods were run under the query-aware setting.

\subsection{TurboQuant Reproduction Details}

TurboQuant \cite{turboquant} randomly rotates each key/value vector so its coordinates become Beta-distributed, then applies per-coordinate optimal scalar quantizers; the Prod variant adds a 1-bit QJL residual transform for unbiased inner-product estimation. We evaluate the 3.5-bit-per-channel configuration, which the paper reports as quality-neutral relative to the full cache. \cite{turboquant} did not release public code at the time of this submission, and therefore we opted to reproduce their work relying on community resources.

Our implementation builds on the main open-source reproduction \citep{turboquant-repo-sero}, with outlier handling matching two other public reference implementations - \cite{turboquant-repo-amesianx}, \cite{turboquant-repo-thetom}. Rather than the channel split described in the original text: per key vector we retain the top-k highest-magnitude channels in fp16 and apply a uniform b-bit base quantizer to the remainder, plus a 128-token fp16 recent window. Concretely, TQ3.5 uses b=3, k=4. We found the fixed high-/low-bit channel split described in the paper text did not reproduce the reported near-lossless quality in our setup ($\sim$7-point LongBench drop vs. full KV), whereas the uniform-base + fp16-outlier scheme used by the reference implementations does. We also point out that the vLLM implementation \citep{turboquant-vllm} also found that deviations from the verbatim method are required to reproduce the reported numbers. They chose to exclude the first two and last two transformer layers from quantization to match the reported results, at the cost of a lower compression rate. Our choice avoids this, compressing all layers.

\section{Experiments - Additional Results}
\label{app:results}
\subsection{Needle in a Haystack}
\label{app:niah}

We follow the passkey retrieval protocol of \citet{kivi2024}. Depth is the fractional position of the passkey in the context, so depth $0$ places it at the beginning and depth $100$ immediately before the query. We sweep context length and needle depth over 10 depths per length and 10 samples per depth, giving 100 samples per context length, and resample the passkey for every sample. A sample is scored 1 if the recovered digit string matches the planted one exactly and 0 otherwise, so each cell of the heatmaps is the mean over its 10 samples.

\begin{figure*}[t]
\centering
\includegraphics[width=\textwidth]{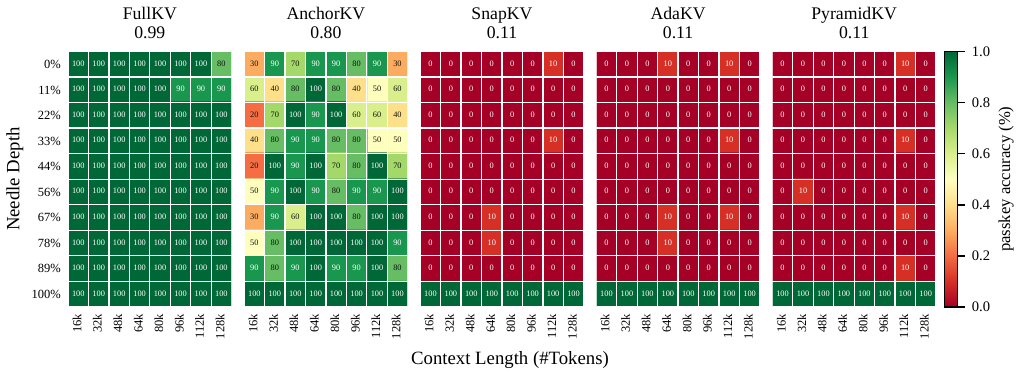}
\caption{Needle in a haystack at $10\times$ compression. Each cell is the mean score over the 10 samples at that context length and needle depth.}
\label{fig:niah-x10-full}
\end{figure*}

Figures~\ref{fig:niah-x5-full} and \ref{fig:niah-x10-full} give the full grids at $5\times$ and $10\times$. AnchorKV recovers the needle across nearly the whole grid at both ratios, while the eviction baselines collapse to almost zero away from the most recent positions. All methods are exact along the bottom row, where the needle falls inside the recency window they store verbatim. That row is the one region where the two families agree, and it marks the limit of what a recency window alone can do: everything above it is retrieval the compressed cache has to support on its own.

\subsection{RULER and LongBench - Full Results}
\begin{figure*}[t]\centering
  \includegraphics[width=0.8\textwidth]{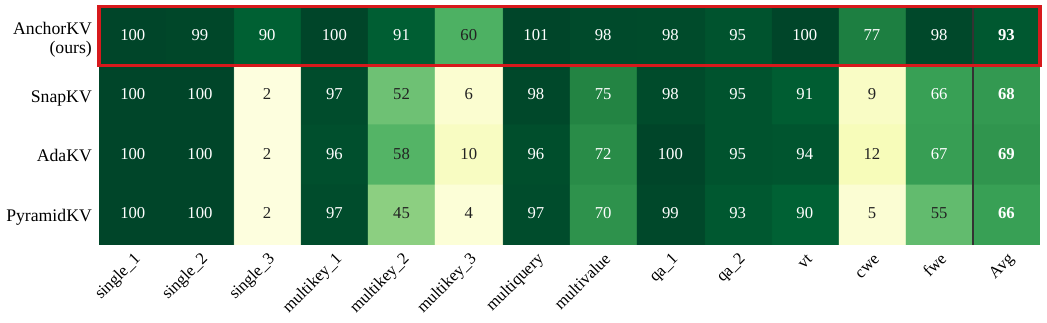}
  \caption{Llama-3.1-8B-Instruct on RULER (32k).}\label{fig:pertask-llama8b_ruler32k}
\end{figure*}

\begin{figure*}[t]\centering
  \includegraphics[width=0.8\textwidth]{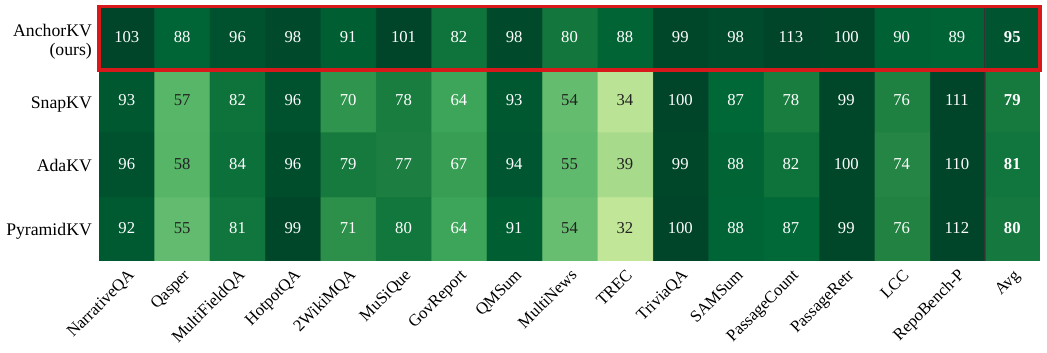}
  \caption{Llama-3.1-8B-Instruct on LongBench.}\label{fig:pertask-llama8b_longbench}
\end{figure*}

\begin{figure*}[t]\centering
  \includegraphics[width=0.8\textwidth]{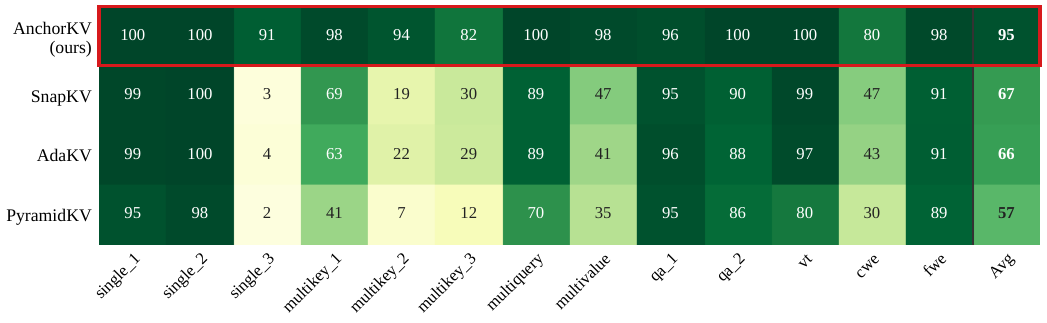}
  \caption{Mistral-Small-3.1-24B-Instruct on RULER (32k).}\label{fig:pertask-mistral24b_ruler32k}
\end{figure*}

\begin{figure*}[t]\centering
  \includegraphics[width=0.8\textwidth]{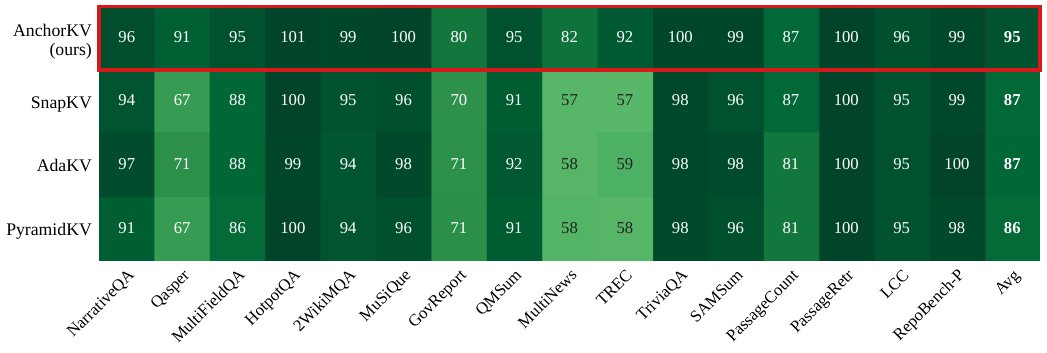}
  \caption{Mistral-Small-3.1-24B-Instruct on LongBench.}\label{fig:pertask-mistral24b_longbench}
\end{figure*}

\begin{figure*}[t]\centering
  \includegraphics[width=0.8\textwidth]{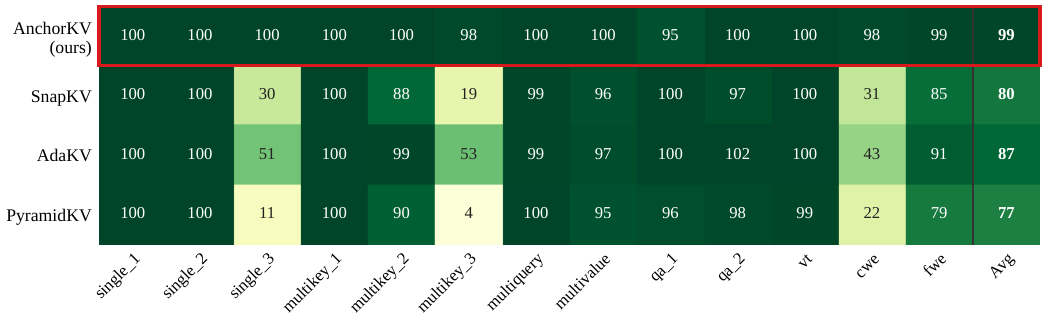}
  \caption{Llama-3.1-70B-Instruct on RULER (32k).}\label{fig:pertask-llama70b_ruler32k}
\end{figure*}

\begin{figure*}[t]\centering
  \includegraphics[width=0.8\textwidth]{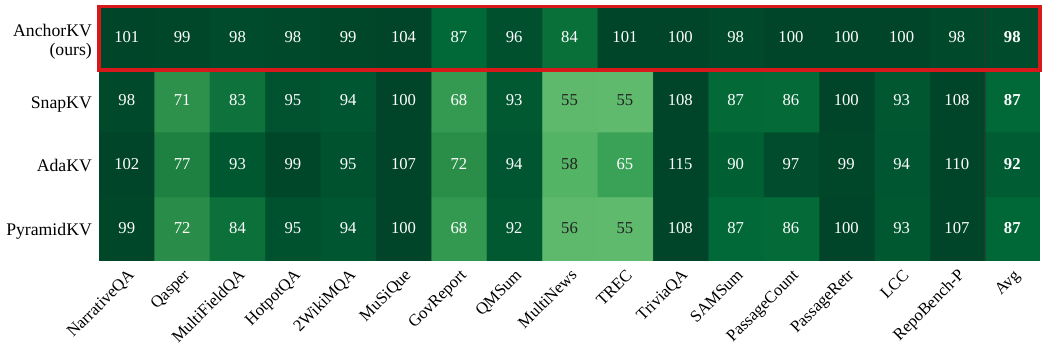}
  \caption{Llama-3.1-70B-Instruct on LongBench.}\label{fig:pertask-llama70b_longbench}
\end{figure*}

We provide tables with the full results on all the subtasks for all of the models and baselines in Tables~\ref{tab:ruler32k-llama8b} - \ref{tab:longbench-llama70b}. Additionally, per-task heatmaps for the three models on RULER-32k and Longbench are displayed in Figures~\ref{fig:pertask-llama8b_ruler32k} - \ref{fig:pertask-llama70b_longbench}.

\section{Additional Ablation Results and Analysis}
\label{app:ablation}

\begin{figure}[!tb]
\centering
\includegraphics[width=0.9\columnwidth]{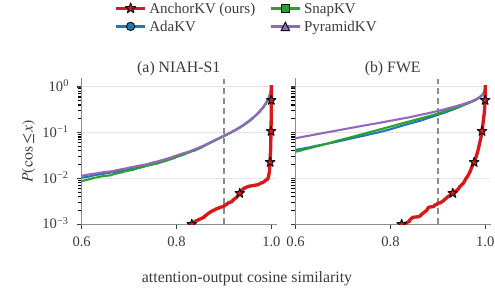}
\caption{Attention-output fidelity at $20\times$ compression, per (layer, head,
sample) cell (Llama-3.1-8B-Instruct, RULER 32K; log $y$-axis, dashed line at
$\cos = 0.9$). Below that threshold AnchorKV leaves $0.2\%$ of cells against
$8.4\%$ for the best baseline on NIAH-S1, and $0.3\%$ against $24\%$ on FWE.}
\label{fig:fidelity}
\end{figure}
\subsection{Attention-Output Fidelity}
\label{app:attn_out}
Figure~\ref{fig:fidelity} measures the quantity the method targets directly, rather than through downstream accuracy: how closely each compressed cache reproduces the attention output of the uncompressed one. For every (layer, head, sample) cell we take the cosine similarity between the two outputs and plot the cumulative distribution on a log axis. A method that reconstructs the output exactly places all of its mass at $\cos = 1$, so its curve stays flat and turns up only at the far right. We mark $\cos = 0.9$ as a reference: mass to the left of that line is cells whose output has moved substantially.

AnchorKV leaves $0.2\%$ of cells below the threshold on NIAH-S1 and $0.3\%$ on FWE. The strongest eviction baseline leaves $8.4\%$ and $24\%$, forty to eighty times as many, and its mass is spread across the whole range rather than gathered near one. The gap is widest on FWE, an aggregation task whose answer draws on many tokens spread through the context. Scoring tokens once and discarding the rest damages a large share of the outputs there, whereas a coarsely stored token still enters the softmax and contributes what its anchor preserves.

\subsection{Where the residual budget goes.}
\label{app:head_residuals}

\begin{figure*}[t]
\centering
\includegraphics[width=0.9\textwidth]{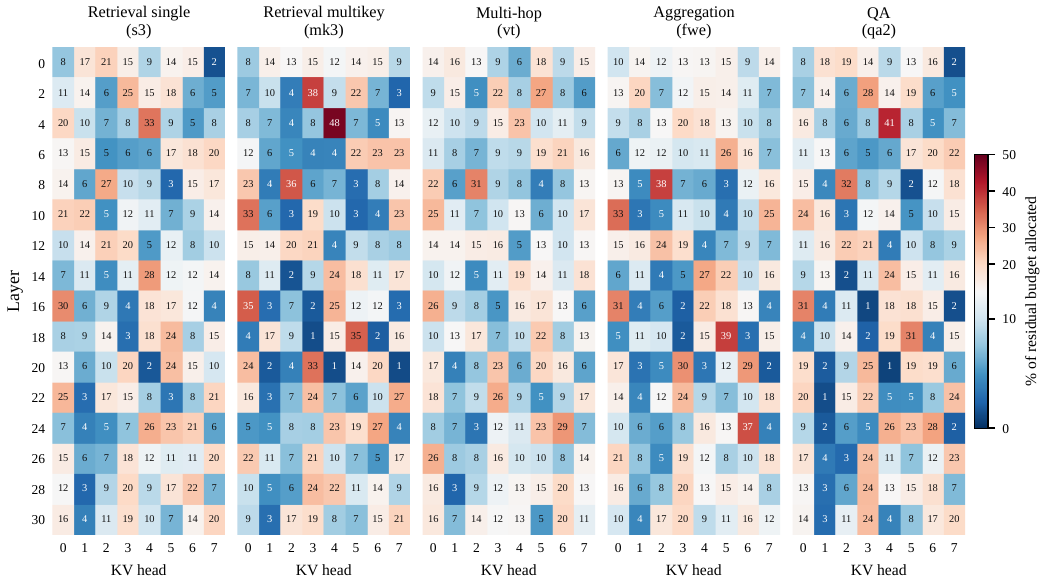}
\caption{Share of each layer's residual budget taken by each KV head, AnchorKV at $20\times$ on Llama-3.1-8B at RULER 32K, averaged over 20 samples per task.
Rows sum to $100\%$, so a per-head split would fix every cell at $12.5\%$. As we can see the head sharing is being used differently across layers as well as across tasks.}
\label{fig:head-budget}
\end{figure*}

Figure~\ref{fig:head-budget} visualizes how AnchorKV distributes each layer's residual budget across the eight KV heads. Each cell gives the share of a layer's residual tokens assigned to one head, so a naive per-head split would leave every cell at the $12.5\%$ reference (white); red cells receive more than this share and blue cells less. The allocation is strongly non-uniform: within a single layer the budget concentrates heavily on a few heads, with individual cells reaching $35$-$48\%$ while others in the same row fall below $5\%$. This concentration is not tied to a fixed set of privileged heads, the head that receives the largest share changes from layer to layer, so no head dominates globally and only a mild preference (head~5) survives once the per-layer structure is averaged away. Together these observations show that the shared-head mechanism is actively exercised rather than decorative: cross-head competition continuously redirects budget toward the heads each layer relies on, which a uniform $12.5\%$ split could not do.

\section{Efficiency and Memory}
\label{app:systems}
 
This appendix expands Section~4.4. All measurements use
Llama-3.1-8B-Instruct in bf16 on one 80\,GiB A100; the baseline is the
uncompressed bf16 cache (FullKV), AnchorKV runs at a $20\times$ target
with the hyperparameters of Appendix~B.3, memory excludes weights, and
every cell is the median of five repeats in a fresh process.
  
% ---------------------------------------------------------------------------
\begin{figure*}[t]
\centering
\includegraphics[width=\textwidth]{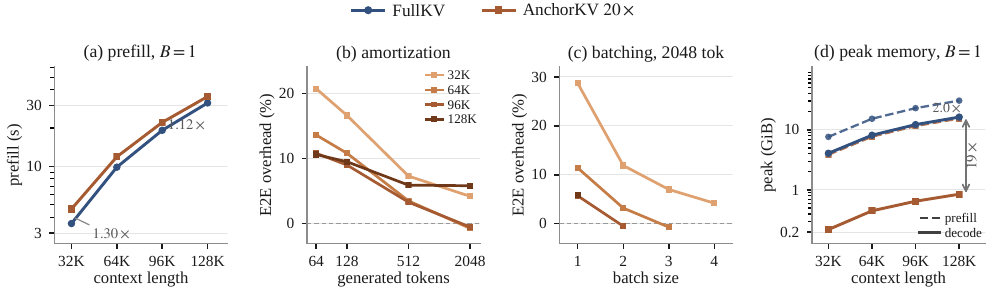}
\caption{Prefill Overhead and Amortization. Same setup as
Figure~6. (a) Prefill wall clock at $B{=}1$: the one-shot
compression pass costs $1.30\times$ at 32K and $1.12\times$ at 128K, and its relative cost shrinks as context grows. (b) End-to-end \texttt{generate()}
overhead against output length, at each context's largest common batch
(32K/$B{=}4$, 64K/$B{=}3$, 96K/$B{=}2$, 128K/$B{=}1$): the pass
amortizes over decode steps, reaching $+4.2\%$ at 32K, $-0.7\%$ at 64K
and $-0.5\%$ at 96K for 2048 tokens. The 128K curve sits above 64K only because memory caps it at
$B{=}1$, where the per-request decode gap is largest. (c) Overhead against batch size at 2048 generated
tokens: batching converges the decode step times, from $+28.7\%$ at
$B{=}1$ to $+4.2\%$ at $B{=}4$ (32K). (d) Peak memory
per request at $B{=}1$, log scale: prefill peaks (dashed) and decode
peaks (solid). Compression cuts the prefill transient $2.0\times$ and
the decode state $19\times$; AnchorKV's prefill peak lands at FullKV's
decode peak.}
\label{fig:prefill}
\end{figure*}

% ---------------------------------------------------------------------------
\begin{table}[!tb]
\centering
\small
\setlength{\tabcolsep}{5pt}
\begin{tabular}{lrrrr}
\toprule
Context & $B$ & FullKV & AnchorKV & ratio \\
\midrule
32K  & 1 & 25.97 & 34.65 & 1.33 \\
32K  & 2 & 36.77 & 37.79 & 1.03 \\
32K  & 4 & 57.79 & \textbf{56.13} & 0.97 \\
64K  & 1 & 36.25 & 37.00 & 1.02 \\
64K  & 2 & 57.65 & \textbf{55.11} & 0.96 \\
96K  & 1 & 46.72 & \textbf{45.55} & 0.97 \\
96K  & 2 & 78.48 & \textbf{72.39} & 0.92 \\
128K & 1 & 57.23 & \textbf{55.82} & 0.98 \\
\bottomrule
\end{tabular}
\caption{Decode latency (ms/token) in every cell where both arms fit.
Bold marks cells where AnchorKV decodes faster; from 64K upward it is at
parity or better at every batch.}
\label{tab:decode-latency}
\end{table}

\subsection{Implementation}
\label{app:eff-kernel}
 
AnchorKV decodes with the \texttt{decode\_attention\_compact} kernel, which reads
the compressed tensors of Table~1 directly and never rebuilds the dense
cache. Anchors and the trailing window are held dense in bf16 with keys
pre-RoPE; every other position stores a nearest-anchor index and one bf16 coefficient $\gamma_t$ (Eq.~\ref{eq:projection}) per side, and the
two-bit residuals are compacted to the tokens that carry one, with a
bitmask resolving position to slot, so a token without a residual costs
one bit and no slot.
 
The execution pattern is motivated by FlashAttention-2 (Dao 2024): the
context is walked in tiles under an online softmax, and nothing dense is
ever written back. Each decode step launches three Triton kernels per
layer. The score kernel reconstructs keys in 16-token tiles entirely in
registers, applies RoPE from the stored positions, contracts the tile
against all four query rows of its GQA group in one tensor-core op, and
emits only fp32 logits: 16 bytes per token instead of a 256-byte
reconstructed key, a $16\times$ traffic reduction. The value kernel never
reconstructs values at all: since the inverse Hadamard transform is
linear, it accumulates the anchor-projection and rotated-residual terms
separately and, at 17.3\% residual density, walks values residual-major,
visiting only the roughly 11 of 64 positions per chunk that store a
residual. The combine kernel reduces the partials, applies the inverse
transform once per head per step rather than once per token, and attends
the exact window (recent prompt tokens plus every generated token, kept
dense) under the same online softmax. The steady-state residency is
therefore the compressed state plus a few MB of workspace, and the
measured decode peak tracks it: $16.26$ against $0.85$\,GiB at 128K, a
$19.1\times$ reduction (Figure~\ref{fig:prefill}d).

\subsection{Generated Tokens}
\label{app:eff-gentok}
 
Tokens produced during decoding are appended to the exact window, dense
in bf16, and are never compressed: the append path is identical to a
standard cache, decoding pays no per-step compression work, and the
compressed prefix and the exact suffix meet in the same online softmax
inside the combine kernel. Because the suffix is dense, the measured
ratio drifts below the operating point as the output grows: the exact
$20\times$ at the moment of compression becomes $19.3\times$ after 60
decode steps at 32K, and a compressed-prompt-plus-dense-suffix model
reproduces the measurement to three decimals, so nothing is lost in the
compressor. The drift is not specific to AnchorKV; any method that
compresses a prompt and then decodes uncompressed has it. We quote the
at-compression figure as the operating point.
 
\subsection{Prefill}
\label{app:eff-prefill}
Compression runs once, per layer, at the end of prefill (Algorithm~1),
and is the only compression work in the request. The pass is built from
batched matrix products and fast Hadamard transforms, remains a small
fraction of prefill compute, and its relative cost shrinks with
context: prefill takes $1.30\times$ the FullKV time at 32K,
$1.15\times$ at 96K and $1.12\times$ at 128K at $B{=}1$
(Figure~\ref{fig:prefill}a), and stays between $+12\%$ and $+30\%$ over
the grid. The prefill peak itself is $2.0\times$ smaller than FullKV's, and
since the prefill peak is what binds the batch, this is what roughly
doubles the concurrency ceiling at every context
(Table~\ref{tab:capacity}); all measurements in this appendix use
unchunked prefill.

% ---------------------------------------------------------------------------
\begin{table}[!tb]
\centering
\small
\setlength{\tabcolsep}{5pt}
\begin{tabular}{lrrrr}
\toprule
 & \multicolumn{2}{c}{FullKV} & \multicolumn{2}{c}{AnchorKV} \\
\cmidrule(lr){2-3} \cmidrule(lr){4-5}
Context & max $B$ & tok/s & max $B$ & tok/s \\
\midrule
32K  & 6 & 76.3 & \textbf{11} & 92.1 \\
64K  & 3 & 38.2 & \textbf{6}  & 48.0 \\
96K  & 2 & 25.5 & \textbf{4}  & 32.2 \\
128K & 1 & 17.5 & \textbf{2}  & 21.8 \\
\bottomrule
\end{tabular}
\caption{Concurrency ceiling on one 80\,GiB A100: the largest batch each
arm can serve, with aggregate throughput there. Every bound is exact
(the largest success and the first failure are consecutive batch
sizes). AnchorKV roughly doubles the ceiling and improves throughput at
each arm's own frontier by 21--26\%.}
\label{tab:capacity}
\end{table}

\subsection{End-to-End Overhead}
\label{app:eff-e2e}
 
Tables~\ref{tab:e2e-headline} and \ref{tab:e2e-full} report the wall
clock of full \texttt{model.generate()} calls, covering prefill,
compression, and all decode steps, over output lengths of 64 to 2048
tokens; positive overhead means AnchorKV is slower. Because the
compression pass is a one-shot cost attached to prefill, the overhead
shrinks along every axis that matters in practice.
 
With output length, the pass amortizes over decode steps: at 32K,
$B{=}4$, the overhead falls from $+20.7\%$ at 64 generated tokens to
$+4.2\%$ at 2048, and at 64K, $B{=}3$, from $+13.6\%$ to $-0.7\%$,
with $-0.5\%$ at 96K, $B{=}2$: the compressed run finishes faster end
to end (Figure~\ref{fig:prefill}b). With batch size, the decode steps of the
two arms converge as the full cache becomes bandwidth-bound while
AnchorKV reads $20\times$ fewer bytes: at 2048 tokens the overhead falls
from $+28.7\%$ at $B{=}1$ to $+4.2\%$ at $B{=}4$
(Figure~\ref{fig:prefill}c), and at 32K, $B{=}4$ the AnchorKV decode
step is already faster (56.1 against 57.8\,ms/token). With context, both
the prefill ratio and the decode gap close, and from 64K upward AnchorKV
decodes at parity or better at every batch
(Table~\ref{tab:decode-latency}). In the regime compression exists for, long
contexts, full batches and non-trivial outputs, the end-to-end price of
keeping every token at $20\times$ compression is under $5\%$ and
occasionally negative, while the decode footprint is $19\times$ smaller
and the concurrency ceiling roughly doubles
(Table~\ref{tab:capacity}).

 % ---------------------------------------------------------------------------
\begin{table}[!tb]
\centering
\small
\setlength{\tabcolsep}{4.5pt}
\begin{tabular}{lrrrrr}
\toprule
 & & \multicolumn{4}{c}{generated tokens} \\
\cmidrule(lr){3-6}
Context & $B$ & 64 & 128 & 512 & 2048 \\
\midrule
32K  & 4 & $+20.7\%$ & $+16.6\%$ & $+7.3\%$ & $+4.2\%$ \\
64K  & 3 & $+13.6\%$ & $+10.8\%$ & $+3.5\%$ & $\mathbf{-0.7\%}$ \\
96K  & 2 & $+10.8\%$ & $+9.0\%$  & $+3.3\%$ & $\mathbf{-0.5\%}$ \\
128K & 1 & $+10.6\%$ & $+9.5\%$  & $+5.9\%$ & $+5.8\%$ \\
\bottomrule
\end{tabular}
\caption{End-to-end \texttt{generate()} overhead of AnchorKV at a
$20\times$ target over FullKV, at each context's largest batch where
both arms fit. Positive means AnchorKV is slower; bold marks the cell
where the compressed run is faster end to end.}
\label{tab:e2e-headline}
\end{table}

\subsection{Prefill-Decode Disaggregation}
\label{app:eff-disagg}
 
The two phases stress memory differently: the prefill peak is the
activation transient, which compression cuts $2.0\times$, while the
decode residency is per-request cache state, which it cuts $19\times$
(Figure~\ref{fig:prefill}d).
Serving stacks that disaggregate prefill from decode split these
cleanly. Prefill nodes absorb the one-shot compression pass at its
$1.12$--$1.30\times$ cost and ship the compressed state, shrinking the
KV transfer by $20\times$; each decode node then holds $19\times$ less
state per request, so a replica that fits one FullKV request at 128K
holds nineteen compressed ones, decoding at parity or better. The
overhead analyzed in this appendix is thus confined to the compute-bound
phase, while the memory-bound phase receives the full reduction.

\begin{table*}[t]
\centering
\small
\setlength{\tabcolsep}{5.5pt}
\begin{tabular}{clrrrrrr}
\toprule
 & & & \multicolumn{2}{c}{Latency (s)} & &
 \multicolumn{2}{c}{Throughput (tok/s)} \\
\cmidrule(lr){4-5} \cmidrule(lr){7-8}
Tokens & Context & $B$ & Full & Ours & Overhead & Full & Ours \\
\midrule
\multirow{10}{*}{64} & 32K & 1 & 5.26 & 6.89 & +30.9\% & 12.16 & 9.29 \\
 & 32K & 2 & 9.52 & 11.52 & +21.0\% & 13.45 & 11.11 \\
 & 32K & 3 & 13.74 & 16.48 & +19.9\% & 13.97 & 11.65 \\
 & 32K & 4 & 17.95 & 21.66 & +20.7\% & 14.26 & 11.82 \\
 & 64K & 1 & 12.22 & 14.32 & +17.2\% & 5.24 & 4.47 \\
 & 64K & 2 & 23.66 & 27.09 & +14.5\% & 5.41 & 4.73 \\
 & 64K & 3 & 35.10 & 39.88 & +13.6\% & 5.47 & 4.81 \\
 & 96K & 1 & 22.20 & 25.08 & +13.0\% & 2.88 & 2.55 \\
 & 96K & 2 & 43.88 & 48.60 & +10.8\% & 2.92 & 2.63 \\
 & 128K & 1 & 35.13 & 38.86 & +10.6\% & 1.82 & 1.65 \\
\midrule
\multirow{10}{*}{128} & 32K & 1 & 7.00 & 9.12 & +30.4\% & 18.30 & 14.04 \\
 & 32K & 2 & 11.94 & 14.05 & +17.7\% & 21.45 & 18.22 \\
 & 32K & 3 & 16.88 & 19.63 & +16.3\% & 22.76 & 19.56 \\
 & 32K & 4 & 21.79 & 25.39 & +16.6\% & 23.50 & 20.16 \\
 & 64K & 1 & 14.64 & 16.83 & +14.9\% & 8.74 & 7.61 \\
 & 64K & 2 & 27.43 & 30.83 & +12.4\% & 9.33 & 8.30 \\
 & 64K & 3 & 40.24 & 44.59 & +10.8\% & 9.54 & 8.61 \\
 & 96K & 1 & 25.28 & 28.09 & +11.1\% & 5.06 & 4.56 \\
 & 96K & 2 & 48.99 & 53.39 & +9.0\% & 5.23 & 4.80 \\
 & 128K & 1 & 38.98 & 42.68 & +9.5\% & 3.28 & 3.00 \\
\midrule
\multirow{10}{*}{512} & 32K & 1 & 17.36 & 22.49 & +29.5\% & 29.48 & 22.76 \\
 & 32K & 2 & 26.49 & 29.37 & +10.9\% & 38.65 & 34.86 \\
 & 32K & 3 & 35.50 & 38.49 & +8.4\% & 43.27 & 39.91 \\
 & 32K & 4 & 44.50 & 47.76 & +7.3\% & 46.02 & 42.88 \\
 & 64K & 1 & 28.97 & 31.85 & +9.9\% & 17.67 & 16.07 \\
 & 64K & 2 & 50.02 & 52.81 & +5.6\% & 20.47 & 19.39 \\
 & 64K & 3 & 70.86 & 73.34 & +3.5\% & 21.68 & 20.94 \\
 & 96K & 1 & 43.64 & 46.34 & +6.2\% & 11.73 & 11.05 \\
 & 96K & 2 & 79.48 & 82.09 & +3.3\% & 12.88 & 12.47 \\
 & 128K & 1 & 61.40 & 65.03 & +5.9\% & 8.34 & 7.87 \\
\midrule
\multirow{10}{*}{2048} & 32K & 1 & 59.27 & 76.26 & +28.7\% & 34.56 & 26.86 \\
 & 32K & 2 & 85.59 & 95.76 & +11.9\% & 47.86 & 42.77 \\
 & 32K & 3 & 111.30 & 119.14 & +7.0\% & 55.20 & 51.57 \\
 & 32K & 4 & 137.21 & 142.93 & +4.2\% & 59.70 & 57.31 \\
 & 64K & 1 & 86.77 & 96.67 & +11.4\% & 23.60 & 21.18 \\
 & 64K & 2 & 141.41 & 145.87 & +3.2\% & 28.97 & 28.08 \\
 & 64K & 3 & 194.70 & 193.35 & \textbf{$-$0.7\%} & 31.56 & 31.78 \\
 & 96K & 1 & 117.44 & 124.14 & +5.7\% & 17.44 & 16.50 \\
 & 96K & 2 & 202.86 & 201.75 & \textbf{$-$0.5\%} & 20.19 & 20.30 \\
 & 128K & 1 & 151.45 & 160.26 & +5.8\% & 13.52 & 12.78 \\
\bottomrule
\end{tabular}
\caption{Full end-to-end generation grid, Llama-3.1-8B-Instruct on one
80\,GiB A100. Ours is AnchorKV at a $20\times$ target; each point is the
wall clock of one \texttt{model.generate()} call, covering prefill,
compression, and all decode steps. Throughput is aggregate over the
batch. Bold marks cells where AnchorKV is faster end to end.}
\label{tab:e2e-full}
\end{table*}

% \newcommand{\pertaskfig}[3]{%
%   \begin{figure*}[t]\centering
%     \includegraphics[width=0.8\textwidth]{ImagesAppendix/#1.pdf}
%     \caption{#2 on #3.}\label{fig:pertask-#1}
%   \end{figure*}}

% \input{Tables/8b_ruler32k}
% \pertaskfig{llama8b_ruler32k}{Llama-3.1-8B-Instruct}{RULER (32k)}
% \pertaskfig{llama8b_longbench}{Llama-3.1-8B-Instruct}{LongBench}
% \pertaskfig{mistral24b_ruler32k}{Mistral-Small-3.1-24B-Instruct}{RULER (32k)}
% \pertaskfig{mistral24b_longbench}{Mistral-Small-3.1-24B-Instruct}{LongBench}
% \pertaskfig{llama70b_ruler32k}{Llama-3.1-70B-Instruct}{RULER (32k)}
% \pertaskfig{llama70b_longbench}{Llama-3.1-70B-Instruct}{LongBench}
% \input{Tables/8b_ruler64k}
% \input{Tables/8b_ruler96k}
% \input{Tables/8b_longbench}
% \input{Tables/24b_ruler32k}
% \input{Tables/24b_ruler64k}
% \input{Tables/24b_longnbench}
% \input{Tables/70b_ruler32k}
% \input{Tables/70b_ruler64k}
% \input{Tables/70b_longbench}

\begin{table*}[t]
\centering
\setlength{\tabcolsep}{5pt}
\small
\begin{tabular}{@{}llrrrrrrrrrrrrrr@{}}
\toprule
Ratio & Method & S1 & S2 & S3 & MK1 & MK2 & MK3 & MQ & MV & VT & CWE & FWE & QA1 & QA2 & Avg \\
\midrule
$1\times$ & FullKV & 100.0 & 100.0 & 100.0 & 99.0 & 100.0 & 99.0 & 97.8 & 99.5 & 99.6 & 27.2 & 88.3 & 81.0 & 56.0 & 88.3 \\
\midrule
$4.5\times$ & TurboQuant & 100.0 & 100.0 & 100.0 & 98.0 & 97.0 & 89.0 & 98.0 & 97.5 & 97.6 & 19.5 & 89.0 & 74.5 & 54.0 & 85.7 \\
\midrule
\multirow{4}{*}{$5\times$} & SnapKV & 100.0 & 97.0 & 29.0 & 98.0 & 95.0 & 76.0 & 98.2 & 96.5 & 97.6 & 13.4 & 72.3 & 83.0 & 56.0 & 77.9 \\
 & AdaKV & 100.0 & 100.0 & 47.0 & 99.0 & 98.0 & 88.0 & 98.0 & 99.5 & 99.6 & 16.5 & 77.7 & 81.0 & 55.0 & 81.5 \\
 & PyramidKV & 100.0 & 99.0 & 15.0 & 98.0 & 89.0 & 71.0 & 96.8 & 99.0 & 98.4 & 6.4 & 65.3 & 81.0 & 55.0 & 74.9 \\
 & AnchorKV & 100.0 & 100.0 & 99.0 & 99.0 & 99.0 & 96.0 & 97.8 & 97.5 & 99.8 & 23.4 & 88.7 & 80.0 & 55.0 & \textbf{87.3} \\
\midrule
\multirow{4}{*}{$10\times$} & SnapKV & 100.0 & 100.0 & 6.0 & 97.0 & 71.0 & 40.0 & 96.8 & 92.5 & 95.6 & 5.4 & 63.3 & 85.0 & 53.0 & 69.7 \\
 & AdaKV & 100.0 & 99.0 & 7.0 & 98.0 & 82.0 & 52.0 & 97.8 & 94.2 & 96.8 & 8.3 & 71.7 & 83.0 & 55.0 & 72.7 \\
 & PyramidKV & 100.0 & 98.0 & 4.0 & 98.0 & 66.0 & 28.0 & 97.8 & 92.2 & 95.6 & 3.0 & 56.0 & 81.0 & 54.0 & 67.2 \\
 & AnchorKV & 100.0 & 100.0 & 99.0 & 99.0 & 99.0 & 90.0 & 97.5 & 97.5 & 99.8 & 24.2 & 89.0 & 80.0 & 53.0 & \textbf{86.8} \\
\midrule
\multirow{4}{*}{$15\times$} & SnapKV & 100.0 & 100.0 & 3.0 & 97.0 & 59.0 & 13.0 & 97.0 & 85.2 & 93.6 & 3.4 & 59.0 & 80.0 & 53.0 & 64.9 \\
 & AdaKV & 100.0 & 100.0 & 2.0 & 96.0 & 67.0 & 27.0 & 95.8 & 85.5 & 95.2 & 5.4 & 63.0 & 83.0 & 53.0 & 67.1 \\
 & PyramidKV & 100.0 & 100.0 & 2.0 & 98.0 & 56.0 & 12.0 & 97.2 & 80.0 & 93.2 & 2.1 & 52.7 & 80.0 & 52.0 & 63.5 \\
 & AnchorKV & 100.0 & 100.0 & 98.0 & 99.0 & 97.0 & 80.0 & 98.2 & 98.0 & 99.8 & 24.1 & 88.0 & 80.0 & 55.0 & \textbf{85.9} \\
\midrule
\multirow{4}{*}{$20\times$} & SnapKV & 100.0 & 100.0 & 2.0 & 96.0 & 52.0 & 6.0 & 96.2 & 74.5 & 90.8 & 2.5 & 58.0 & 79.0 & 53.0 & 62.3 \\
 & AdaKV & 100.0 & 100.0 & 2.0 & 95.0 & 58.0 & 10.0 & 94.0 & 71.8 & 93.4 & 3.3 & 59.0 & 81.0 & 53.0 & 63.1 \\
 & PyramidKV & 100.0 & 100.0 & 2.0 & 96.0 & 45.0 & 4.0 & 94.8 & 69.8 & 89.4 & 1.4 & 48.3 & 80.0 & 52.0 & 60.2 \\
 & AnchorKV & 100.0 & 99.0 & 90.0 & 99.0 & 91.0 & 59.0 & 98.2 & 97.5 & 99.8 & 20.9 & 87.0 & 79.0 & 53.0 & \textbf{82.6} \\
\bottomrule
\end{tabular}
\caption{RULER-32K on Llama-3.1-8B-Instruct.}
\label{tab:ruler32k-llama8b}
\end{table*}

\begin{table*}[t]
\centering
\setlength{\tabcolsep}{5pt}
\small
\begin{tabular}{@{}llrrrrrrrrrrrrrr@{}}
\toprule
Ratio & Method & S1 & S2 & S3 & MK1 & MK2 & MK3 & MQ & MV & VT & CWE & FWE & QA1 & QA2 & Avg \\
\midrule
$1\times$ & FullKV & 100.0 & 100.0 & 100.0 & 100.0 & 99.0 & 97.0 & 98.0 & 97.5 & 97.2 & 6.1 & 88.0 & 84.0 & 56.0 & 86.4 \\
\midrule
$4.5\times$ & TurboQuant & 100.0 & 100.0 & 99.0 & 100.0 & 96.0 & 64.0 & 98.0 & 95.5 & 90.8 & 3.9 & 87.7 & 75.2 & 54.0 & 81.8 \\
\midrule
\multirow{4}{*}{$5\times$} & SnapKV & 100.0 & 98.0 & 56.0 & 100.0 & 84.0 & 69.0 & 98.2 & 92.8 & 96.2 & 1.1 & 77.7 & 84.0 & 54.0 & 77.8 \\
 & AdaKV & 100.0 & 99.0 & 80.0 & 100.0 & 94.0 & 76.0 & 99.0 & 97.0 & 97.8 & 2.2 & 81.0 & 83.0 & 57.0 & 82.0 \\
 & PyramidKV & 100.0 & 99.0 & 37.0 & 99.0 & 76.0 & 64.0 & 98.2 & 92.2 & 97.0 & 1.7 & 75.0 & 80.0 & 56.0 & 75.0 \\
 & AnchorKV & 100.0 & 100.0 & 100.0 & 100.0 & 98.0 & 90.0 & 97.5 & 95.5 & 97.2 & 5.6 & 87.0 & 83.0 & 57.0 & \textbf{85.4} \\
\midrule
\multirow{4}{*}{$10\times$} & SnapKV & 100.0 & 96.0 & 13.0 & 99.0 & 68.0 & 39.0 & 96.0 & 81.8 & 93.8 & 0.5 & 73.7 & 79.0 & 54.0 & 68.7 \\
 & AdaKV & 100.0 & 99.0 & 28.0 & 99.0 & 77.0 & 53.0 & 98.8 & 91.8 & 95.0 & 1.1 & 76.7 & 82.0 & 55.0 & 73.6 \\
 & PyramidKV & 100.0 & 98.0 & 7.0 & 99.0 & 59.0 & 36.0 & 98.5 & 84.2 & 94.8 & 0.7 & 70.0 & 79.0 & 55.0 & 67.8 \\
 & AnchorKV & 100.0 & 100.0 & 100.0 & 100.0 & 96.0 & 88.0 & 97.5 & 95.2 & 97.2 & 5.4 & 86.3 & 84.0 & 55.0 & \textbf{85.0} \\
\midrule
\multirow{4}{*}{$15\times$} & SnapKV & 100.0 & 95.0 & 4.0 & 99.0 & 49.0 & 23.0 & 95.8 & 74.8 & 90.8 & 0.6 & 72.0 & 81.0 & 54.0 & 64.5 \\
 & AdaKV & 100.0 & 96.0 & 12.0 & 100.0 & 65.0 & 32.0 & 97.5 & 81.5 & 91.6 & 0.5 & 73.7 & 83.0 & 56.0 & 68.4 \\
 & PyramidKV & 100.0 & 96.0 & 3.0 & 99.0 & 44.0 & 19.0 & 97.5 & 76.2 & 93.0 & 0.5 & 68.7 & 80.0 & 54.0 & 63.9 \\
 & AnchorKV & 100.0 & 100.0 & 100.0 & 100.0 & 97.0 & 77.0 & 97.2 & 95.2 & 97.2 & 6.6 & 86.3 & 81.0 & 54.0 & \textbf{84.0} \\
\midrule
\multirow{4}{*}{$20\times$} & SnapKV & 100.0 & 95.0 & 3.0 & 97.0 & 46.0 & 14.0 & 94.2 & 71.0 & 85.2 & 0.6 & 70.3 & 81.0 & 54.0 & 62.4 \\
 & AdaKV & 100.0 & 96.0 & 7.0 & 100.0 & 52.0 & 20.0 & 94.8 & 73.8 & 88.4 & 0.4 & 70.0 & 83.0 & 55.0 & 64.6 \\
 & PyramidKV & 100.0 & 95.0 & 2.0 & 98.0 & 35.0 & 15.0 & 95.0 & 67.0 & 90.8 & 0.5 & 67.0 & 80.0 & 56.0 & 61.6 \\
 & AnchorKV & 100.0 & 100.0 & 94.0 & 99.0 & 89.0 & 59.0 & 97.0 & 94.5 & 97.2 & 5.1 & 85.7 & 82.0 & 54.0 & \textbf{81.3} \\
\bottomrule
\end{tabular}
\caption{RULER-64K on Llama-3.1-8B-Instruct.}
\label{tab:ruler64k-llama8b}
\end{table*}
\begin{table*}[t]
\centering
\setlength{\tabcolsep}{5pt}
\small
\begin{tabular}{@{}llrrrrrrrrrrrrrr@{}}
\toprule
Ratio & Method & S1 & S2 & S3 & MK1 & MK2 & MK3 & MQ & MV & VT & CWE & FWE & QA1 & QA2 & Avg \\
\midrule
$1\times$ & FullKV & 100.0 & 100.0 & 100.0 & 99.0 & 97.0 & 95.0 & 99.5 & 98.5 & 82.2 & 0.3 & 88.7 & 86.0 & 47.0 & 84.1 \\
\midrule
$4.5\times$ & TurboQuant & 100.0 & 99.0 & 98.0 & 97.0 & 90.0 & 32.0 & 98.2 & 94.2 & 76.6 & 1.0 & 85.7 & 72.9 & 44.0 & 76.1 \\
\midrule
\multirow{4}{*}{$5\times$} & SnapKV & 100.0 & 99.0 & 84.0 & 99.0 & 78.0 & 54.0 & 99.8 & 91.2 & 77.8 & 0.0 & 79.7 & 85.0 & 46.0 & 76.4 \\
 & AdaKV & 100.0 & 100.0 & 91.0 & 99.0 & 92.0 & 82.0 & 100.0 & 97.8 & 77.0 & 0.0 & 82.7 & 86.0 & 47.0 & 81.1 \\
 & PyramidKV & 100.0 & 99.0 & 61.0 & 99.0 & 70.0 & 55.0 & 100.0 & 96.2 & 75.4 & 0.0 & 75.7 & 84.0 & 46.0 & 73.9 \\
 & AnchorKV & 100.0 & 100.0 & 100.0 & 99.0 & 94.0 & 71.0 & 99.5 & 97.5 & 84.4 & 0.5 & 87.0 & 85.0 & 44.0 & \textbf{81.7} \\
\midrule
\multirow{4}{*}{$10\times$} & SnapKV & 100.0 & 100.0 & 44.0 & 98.0 & 51.0 & 23.0 & 99.5 & 86.2 & 75.8 & 0.0 & 75.0 & 83.0 & 46.0 & 67.8 \\
 & AdaKV & 100.0 & 99.0 & 60.0 & 99.0 & 62.0 & 42.0 & 99.2 & 90.5 & 77.0 & 0.0 & 77.7 & 85.0 & 46.0 & 72.1 \\
 & PyramidKV & 100.0 & 99.0 & 23.0 & 98.0 & 40.0 & 24.0 & 99.0 & 88.8 & 77.8 & 0.0 & 66.3 & 84.0 & 45.0 & 65.0 \\
 & AnchorKV & 100.0 & 100.0 & 99.0 & 99.0 & 93.0 & 72.0 & 99.5 & 97.5 & 85.0 & 0.5 & 87.0 & 85.0 & 45.0 & \textbf{81.7} \\
\midrule
\multirow{4}{*}{$15\times$} & SnapKV & 100.0 & 100.0 & 20.0 & 98.0 & 34.0 & 11.0 & 98.5 & 78.5 & 74.8 & 0.0 & 67.3 & 83.0 & 45.0 & 62.3 \\
 & AdaKV & 100.0 & 100.0 & 23.0 & 98.0 & 47.0 & 24.0 & 99.5 & 85.8 & 74.8 & 0.0 & 74.7 & 84.0 & 46.0 & 65.9 \\
 & PyramidKV & 100.0 & 100.0 & 5.0 & 99.0 & 21.0 & 15.0 & 99.0 & 79.0 & 76.0 & 0.0 & 60.3 & 82.0 & 41.0 & 59.8 \\
 & AnchorKV & 100.0 & 100.0 & 96.0 & 99.0 & 93.0 & 52.0 & 99.8 & 96.2 & 85.0 & 0.7 & 87.3 & 83.0 & 45.0 & \textbf{79.8} \\
\midrule
\multirow{4}{*}{$20\times$} & SnapKV & 100.0 & 100.0 & 3.0 & 99.0 & 25.0 & 6.0 & 98.2 & 69.2 & 73.0 & 0.0 & 59.0 & 82.0 & 45.0 & 58.4 \\
 & AdaKV & 100.0 & 100.0 & 9.0 & 98.0 & 34.0 & 12.0 & 98.0 & 76.5 & 74.0 & 0.0 & 69.3 & 83.0 & 46.0 & 61.5 \\
 & PyramidKV & 100.0 & 100.0 & 2.0 & 99.0 & 16.0 & 8.0 & 98.0 & 71.0 & 74.8 & 0.1 & 56.0 & 81.0 & 42.0 & 57.5 \\
 & AnchorKV & 100.0 & 100.0 & 89.0 & 99.0 & 83.0 & 43.0 & 99.8 & 95.8 & 83.8 & 0.3 & 87.0 & 83.0 & 47.0 & \textbf{77.7} \\
\bottomrule
\end{tabular}
\caption{RULER-96K on Llama-3.1-8B-Instruct.}
\label{tab:ruler96k-llama8b}
\end{table*}
\begin{table*}[t]
\centering
\setlength{\tabcolsep}{1.5pt}
\small
\begin{tabular}{@{}llrrrrrrrrrrrrrrrrr@{}}
\toprule
 &  & \multicolumn{3}{c}{Single-Doc QA} & \multicolumn{3}{c}{Multi-Doc QA} & \multicolumn{3}{c}{Summarization} & \multicolumn{3}{c}{Few-shot} & \multicolumn{2}{c}{Synthetic} & \multicolumn{2}{c}{Code} &  \\
\cmidrule(lr){3-5}\cmidrule(lr){6-8}\cmidrule(lr){9-11}\cmidrule(lr){12-14}\cmidrule(lr){15-16}\cmidrule(lr){17-18}
Ratio & Method & NQA & Qsp & MFQA & HQA & 2Wik & MSQ & GovR & QMS & MNws & TREC & TQA & SAMS & PCnt & PRe & LCC & RB-P & Avg \\
\midrule
$1\times$ & FullKV & 31.5 & 48.4 & 56.2 & 58.5 & 52.2 & 35.0 & 35.3 & 25.2 & 27.1 & 73.0 & 90.8 & 43.8 & 10.9 & 100.0 & 65.3 & 46.7 & 50.0 \\
\midrule
$4.2\times$ & TurboQuant & 31.2 & 46.7 & 56.1 & 59.1 & 52.1 & 35.4 & 35.2 & 25.1 & 27.2 & 73.0 & 89.6 & 43.2 & 12.5 & 99.5 & 64.9 & 44.7 & 49.7 \\
\midrule
\multirow{4}{*}{$5\times$} & SnapKV & 30.4 & 44.2 & 54.6 & 58.8 & 49.2 & 33.0 & 28.7 & 24.5 & 22.7 & 40.5 & 91.7 & 40.6 & 13.1 & 99.5 & 53.5 & 48.5 & 45.8 \\
 & AdaKV & 30.1 & 43.1 & 55.4 & 58.6 & 50.1 & 32.8 & 28.6 & 24.4 & 23.3 & 37.0 & 91.9 & 40.1 & 10.6 & 99.5 & 50.5 & 46.2 & 45.1 \\
 & PyramidKV & 30.2 & 43.9 & 55.2 & 58.4 & 49.2 & 30.8 & 28.2 & 24.5 & 22.6 & 39.0 & 91.9 & 39.4 & 11.5 & 99.5 & 53.4 & 47.3 & 45.3 \\
 & AnchorKV & 31.8 & 46.5 & 55.0 & 58.0 & 48.5 & 34.1 & 34.1 & 24.4 & 26.7 & 73.0 & 89.5 & 43.6 & 12.3 & 99.5 & 63.2 & 46.0 & \textbf{49.1} \\
\midrule
\multirow{4}{*}{$10\times$} & SnapKV & 30.9 & 36.9 & 53.6 & 57.0 & 44.4 & 27.9 & 25.8 & 23.7 & 19.8 & 35.5 & 91.1 & 40.1 & 10.6 & 99.5 & 51.0 & 50.4 & 43.6 \\
 & AdaKV & 30.8 & 38.4 & 53.9 & 57.9 & 46.6 & 31.3 & 26.1 & 24.2 & 20.5 & 34.5 & 90.9 & 40.0 & 10.1 & 99.5 & 49.3 & 48.9 & 43.9 \\
 & PyramidKV & 31.1 & 37.6 & 53.1 & 56.2 & 44.1 & 28.3 & 25.5 & 23.9 & 19.7 & 34.5 & 91.1 & 40.2 & 9.6 & 99.5 & 52.0 & 50.1 & 43.5 \\
 & AnchorKV & 32.0 & 44.5 & 54.6 & 57.1 & 50.3 & 34.7 & 32.6 & 24.8 & 25.7 & 72.0 & 89.3 & 43.9 & 13.3 & 99.5 & 60.5 & 43.8 & \textbf{48.7} \\
\midrule
\multirow{4}{*}{$15\times$} & SnapKV & 30.5 & 32.1 & 51.2 & 57.9 & 44.0 & 28.6 & 24.1 & 24.0 & 16.8 & 29.0 & 89.5 & 39.2 & 10.0 & 99.5 & 50.7 & 51.5 & 42.4 \\
 & AdaKV & 30.7 & 33.7 & 50.4 & 57.2 & 43.5 & 29.0 & 24.3 & 23.4 & 17.6 & 33.0 & 90.2 & 39.0 & 10.1 & 99.5 & 48.4 & 49.4 & 42.5 \\
 & PyramidKV & 29.9 & 31.6 & 51.6 & 58.0 & 43.7 & 26.8 & 24.0 & 23.8 & 16.9 & 27.5 & 90.4 & 39.4 & 11.0 & 99.5 & 50.6 & 51.7 & 42.3 \\
 & AnchorKV & 31.3 & 43.6 & 53.3 & 57.3 & 48.7 & 34.7 & 30.8 & 24.4 & 23.9 & 69.0 & 89.5 & 43.4 & 11.8 & 99.5 & 59.5 & 42.3 & \textbf{47.7} \\
\midrule
\multirow{4}{*}{$20\times$} & SnapKV & 29.3 & 27.6 & 46.3 & 55.9 & 36.3 & 27.2 & 22.6 & 23.4 & 14.7 & 24.5 & 90.7 & 38.3 & 8.5 & 99.0 & 49.8 & 51.7 & 40.4 \\
 & AdaKV & 30.2 & 27.9 & 47.0 & 56.5 & 41.5 & 27.0 & 23.5 & 23.6 & 15.0 & 28.5 & 90.0 & 38.7 & 9.0 & 99.5 & 48.7 & 51.2 & 41.1 \\
 & PyramidKV & 29.0 & 26.6 & 45.6 & 57.9 & 36.9 & 28.1 & 22.6 & 22.9 & 14.5 & 23.0 & 91.1 & 38.6 & 9.5 & 99.0 & 49.5 & 52.1 & 40.4 \\
 & AnchorKV & 32.5 & 42.8 & 53.8 & 57.4 & 47.4 & 35.3 & 28.9 & 24.7 & 21.8 & 64.5 & 89.8 & 42.8 & 12.3 & 99.5 & 58.7 & 41.4 & \textbf{47.1} \\
\bottomrule
\end{tabular}
\caption{LongBench on Llama-3.1-8B-Instruct.}
\label{tab:longbench-llama8b}
\end{table*}
\begin{table*}[t]
\centering
\setlength{\tabcolsep}{4pt}
\small
\begin{tabular}{@{}llrrrrrrrrrrrrrr@{}}
\toprule
Ratio & Method & S1 & S2 & S3 & MK1 & MK2 & MK3 & MQ & MV & VT & CWE & FWE & QA1 & QA2 & Avg \\
\midrule
$1\times$ & FullKV & 100.0 & 100.0 & 100.0 & 99.0 & 100.0 & 98.0 & 99.8 & 99.8 & 100.0 & 82.1 & 95.0 & 85.0 & 69.0 & 94.4 \\
\midrule
$4.5\times$ & TurboQuant & 100.0 & 100.0 & 100.0 & 99.0 & 98.0 & 98.0 & 98.5 & 99.8 & 100.0 & 77.1 & 95.3 & 74.2 & 67.0 & 92.8 \\
\midrule
\multirow{4}{*}{$10\times$} & SnapKV & 100.0 & 100.0 & 13.0 & 82.0 & 42.0 & 56.0 & 95.8 & 66.5 & 99.8 & 53.5 & 88.7 & 84.0 & 63.0 & 72.6 \\
 & AdaKV & 100.0 & 100.0 & 14.0 & 80.0 & 46.0 & 54.0 & 96.0 & 68.2 & 100.0 & 47.4 & 89.7 & 83.0 & 63.0 & 72.4 \\
 & PyramidKV & 98.0 & 100.0 & 3.0 & 58.0 & 17.0 & 33.0 & 87.5 & 52.2 & 95.6 & 31.2 & 88.3 & 78.0 & 61.0 & 61.8 \\
 & AnchorKV & 100.0 & 100.0 & 98.0 & 100.0 & 99.0 & 97.0 & 100.0 & 99.8 & 100.0 & 75.8 & 94.7 & 85.0 & 73.0 & \textbf{94.0} \\
\midrule
\multirow{4}{*}{$20\times$} & SnapKV & 99.0 & 100.0 & 3.0 & 68.0 & 19.0 & 29.0 & 89.2 & 47.0 & 98.6 & 38.4 & 86.0 & 81.0 & 62.0 & 63.1 \\
 & AdaKV & 99.0 & 100.0 & 4.0 & 62.0 & 22.0 & 28.0 & 89.2 & 40.8 & 97.4 & 35.3 & 86.3 & 82.0 & 61.0 & 62.1 \\
 & PyramidKV & 95.0 & 98.0 & 2.0 & 41.0 & 7.0 & 12.0 & 70.2 & 34.5 & 79.8 & 25.0 & 84.3 & 81.0 & 59.0 & 53.0 \\
 & AnchorKV & 100.0 & 100.0 & 91.0 & 97.0 & 94.0 & 80.0 & 99.8 & 97.8 & 99.8 & 66.0 & 93.3 & 82.0 & 69.0 & \textbf{90.0} \\
\bottomrule
\end{tabular}
\caption{RULER at 32K on Mistral-Small-3.1-24B-Instruct.}
\label{tab:ruler32k-mistral24b}
\end{table*}
\begin{table*}[t]
\centering
\setlength{\tabcolsep}{5pt}
\small
\begin{tabular}{@{}llrrrrrrrrrrrrrr@{}}
\toprule
Ratio & Method & S1 & S2 & S3 & MK1 & MK2 & MK3 & MQ & MV & VT & CWE & FWE & QA1 & QA2 & Avg \\
\midrule
$1\times$ & FullKV & 99.0 & 100.0 & 100.0 & 97.0 & 96.0 & 96.0 & 97.8 & 99.8 & 99.8 & 49.7 & 94.0 & 83.0 & 61.0 & 90.2 \\
\midrule
$4.5\times$ & TurboQuant & 98.0 & 100.0 & 100.0 & 94.0 & 93.0 & 94.0 & 97.0 & 99.5 & 99.2 & 0.0 & 94.7 & 70.6 & 59.0 & 84.5 \\
\midrule
\multirow{4}{*}{$10\times$} & SnapKV & 100.0 & 100.0 & 7.0 & 86.0 & 56.0 & 20.0 & 95.0 & 73.8 & 96.2 & 35.9 & 84.7 & 78.0 & 54.0 & 68.2 \\
 & AdaKV & 100.0 & 100.0 & 9.0 & 87.0 & 58.0 & 18.0 & 96.2 & 72.2 & 95.2 & 29.4 & 85.3 & 77.0 & 54.0 & 67.8 \\
 & PyramidKV & 100.0 & 98.0 & 3.0 & 67.0 & 35.0 & 10.0 & 89.0 & 59.5 & 93.8 & 20.1 & 85.0 & 77.0 & 54.0 & 60.9 \\
 & AnchorKV & 99.0 & 100.0 & 100.0 & 92.0 & 93.0 & 89.0 & 98.0 & 99.5 & 100.0 & 42.8 & 93.3 & 78.0 & 60.0 & \textbf{88.0} \\
\midrule
\multirow{4}{*}{$20\times$} & SnapKV & 100.0 & 100.0 & 3.0 & 73.0 & 35.0 & 10.0 & 85.5 & 50.5 & 91.4 & 22.6 & 81.7 & 76.0 & 54.0 & 60.2 \\
 & AdaKV & 100.0 & 99.0 & 3.0 & 74.0 & 35.0 & 9.0 & 83.5 & 50.0 & 89.8 & 20.1 & 81.7 & 75.0 & 55.0 & 59.6 \\
 & PyramidKV & 100.0 & 95.0 & 2.0 & 40.0 & 11.0 & 4.0 & 69.8 & 35.2 & 82.4 & 9.4 & 78.7 & 74.0 & 51.0 & 50.2 \\
 & AnchorKV & 99.0 & 100.0 & 67.0 & 92.0 & 79.0 & 57.0 & 97.2 & 97.2 & 99.6 & 32.3 & 94.0 & 77.0 & 60.0 & \textbf{80.9} \\
\bottomrule
\end{tabular}
\caption{RULER at 64K on Mistral-Small-3.1-24B-Instruct.}
\label{tab:ruler64k-mistral24b}
\end{table*}
\begin{table*}[t]
\centering
\setlength{\tabcolsep}{1.5pt}
\small
\begin{tabular}{@{}llrrrrrrrrrrrrrrrrr@{}}
\toprule
 &  & \multicolumn{3}{c}{Single-Doc QA} & \multicolumn{3}{c}{Multi-Doc QA} & \multicolumn{3}{c}{Summarization} & \multicolumn{3}{c}{Few-shot} & \multicolumn{2}{c}{Synthetic} & \multicolumn{2}{c}{Code} &  \\
\cmidrule(lr){3-5}\cmidrule(lr){6-8}\cmidrule(lr){9-11}\cmidrule(lr){12-14}\cmidrule(lr){15-16}\cmidrule(lr){17-18}
Ratio & Method & NQA & Qsp & MFQA & HQA & 2Wik & MSQ & GovR & QMS & MNws & TREC & TQA & SAMS & PCnt & PRe & LCC & RB-P & Avg \\
\midrule
$1\times$ & FullKV & 36.7 & 51.5 & 57.0 & 66.8 & 67.3 & 48.5 & 33.3 & 24.9 & 25.0 & 76.5 & 93.2 & 50.5 & 15.5 & 100.0 & 70.7 & 74.2 & 55.7 \\
\midrule
$4.3\times$ & TurboQuant & 36.9 & 50.2 & 55.3 & 66.7 & 67.9 & 48.9 & 32.6 & 24.6 & 24.3 & 76.5 & 92.6 & 50.0 & 16.0 & 100.0 & 70.2 & 74.1 & 55.4 \\
\midrule
\multirow{4}{*}{$10\times$} & SnapKV & 35.5 & 42.2 & 55.4 & 66.7 & 67.2 & 47.2 & 26.1 & 23.8 & 18.7 & 54.5 & 91.7 & 49.3 & 13.5 & 100.0 & 64.2 & 73.6 & 51.9 \\
 & AdaKV & 35.5 & 44.2 & 55.4 & 66.0 & 66.9 & 47.2 & 25.9 & 23.6 & 19.3 & 56.0 & 91.4 & 50.0 & 15.5 & 100.0 & 65.6 & 73.5 & 52.2 \\
 & PyramidKV & 36.0 & 41.9 & 55.5 & 65.3 & 67.3 & 47.1 & 26.1 & 23.5 & 18.7 & 54.5 & 92.0 & 49.7 & 13.0 & 100.0 & 64.2 & 73.0 & 51.7 \\
 & AnchorKV & 36.1 & 49.7 & 56.2 & 67.3 & 66.8 & 47.3 & 31.2 & 23.9 & 23.6 & 75.0 & 92.7 & 50.5 & 13.5 & 100.0 & 68.4 & 73.2 & \textbf{54.7} \\
\midrule
\multirow{4}{*}{$20\times$} & SnapKV & 34.7 & 34.4 & 50.0 & 66.8 & 63.8 & 46.4 & 23.4 & 22.7 & 14.4 & 43.5 & 91.6 & 48.3 & 13.5 & 100.0 & 67.0 & 73.2 & 49.6 \\
 & AdaKV & 35.7 & 36.5 & 50.0 & 66.1 & 62.9 & 47.5 & 23.6 & 23.0 & 14.4 & 45.5 & 91.7 & 49.3 & 12.5 & 100.0 & 67.3 & 74.0 & 50.0 \\
 & PyramidKV & 33.3 & 34.5 & 49.2 & 66.7 & 63.1 & 46.4 & 23.7 & 22.7 & 14.5 & 44.0 & 91.6 & 48.5 & 12.5 & 100.0 & 67.1 & 73.0 & 49.4 \\
 & AnchorKV & 35.4 & 46.6 & 54.4 & 67.8 & 66.5 & 48.6 & 26.7 & 23.8 & 20.6 & 70.5 & 92.7 & 50.0 & 13.5 & 100.0 & 67.8 & 73.2 & \textbf{53.6} \\
\bottomrule
\end{tabular}
\caption{LongBench on Mistral-Small-3.1-24B-Instruct.}
\label{tab:longbench-mistral24b}
\end{table*}
\begin{table*}[t]
\centering
\setlength{\tabcolsep}{4pt}
\small
\begin{tabular}{@{}llrrrrrrrrrrrrrr@{}}
\toprule
Ratio & Method & S1 & S2 & S3 & MK1 & MK2 & MK3 & MQ & MV & VT & CWE & FWE & QA1 & QA2 & Avg \\
\midrule
$1\times$ & FullKV & 100.0 & 100.0 & 100.0 & 100.0 & 99.0 & 100.0 & 99.8 & 99.8 & 100.0 & 93.5 & 98.3 & 84.0 & 61.0 & 95.0 \\
\midrule
$4.5\times$ & TurboQuant & 100.0 & 100.0 & 100.0 & 100.0 & 99.0 & 98.0 & 100.0 & 99.0 & 100.0 & 91.9 & 98.0 & 74.9 & 59.0 & 93.8 \\
\midrule
\multirow{4}{*}{$10\times$} & SnapKV & 100.0 & 100.0 & 88.0 & 100.0 & 98.0 & 59.0 & 99.8 & 98.5 & 100.0 & 49.7 & 90.3 & 83.0 & 60.0 & 86.6 \\
 & AdaKV & 100.0 & 100.0 & 97.0 & 100.0 & 99.0 & 86.0 & 99.5 & 99.0 & 100.0 & 66.1 & 93.0 & 84.0 & 60.0 & 91.0 \\
 & PyramidKV & 100.0 & 100.0 & 61.0 & 100.0 & 97.0 & 36.0 & 99.8 & 96.8 & 99.8 & 35.1 & 88.7 & 82.0 & 60.0 & 81.2 \\
 & AnchorKV & 100.0 & 100.0 & 100.0 & 100.0 & 99.0 & 100.0 & 99.8 & 99.8 & 100.0 & 92.6 & 98.0 & 80.0 & 61.0 & \textbf{94.6} \\
\midrule
\multirow{4}{*}{$20\times$} & SnapKV & 100.0 & 100.0 & 30.0 & 100.0 & 87.0 & 19.0 & 99.0 & 95.8 & 99.8 & 29.4 & 83.3 & 84.0 & 59.0 & 75.9 \\
 & AdaKV & 100.0 & 100.0 & 51.0 & 100.0 & 98.0 & 53.0 & 98.8 & 96.5 & 100.0 & 39.9 & 89.7 & 84.0 & 62.0 & 82.5 \\
 & PyramidKV & 100.0 & 100.0 & 11.0 & 100.0 & 89.0 & 4.0 & 99.8 & 94.8 & 99.4 & 21.0 & 78.0 & 81.0 & 60.0 & 72.1 \\
 & AnchorKV & 100.0 & 100.0 & 100.0 & 100.0 & 99.0 & 98.0 & 99.8 & 99.8 & 100.0 & 91.8 & 97.3 & 80.0 & 61.0 & \textbf{94.4} \\
\bottomrule
\end{tabular}
\caption{RULER at 32K on Llama-3.1-70B-Instruct.}
\label{tab:ruler32k-llama70b}
\end{table*}
\begin{table*}[t]
\centering
\setlength{\tabcolsep}{4pt}
\small
\begin{tabular}{@{}llrrrrrrrrrrrrrr@{}}
\toprule
Ratio & Method & S1 & S2 & S3 & MK1 & MK2 & MK3 & MQ & MV & VT & CWE & FWE & QA1 & QA2 & Avg \\
\midrule
$1\times$ & FullKV & 100.0 & 100.0 & 100.0 & 100.0 & 99.0 & 99.0 & 99.8 & 95.0 & 100.0 & 22.0 & 94.0 & 79.0 & 59.0 & 88.2 \\
\midrule
$4.5\times$ & TurboQuant & 100.0 & 100.0 & 100.0 & 99.0 & 94.0 & 95.0 & 100.0 & 97.5 & 100.0 & 25.0 & 96.0 & 68.6 & 54.0 & 86.9 \\
\midrule
\multirow{4}{*}{$10\times$} & SnapKV & 100.0 & 100.0 & 74.0 & 100.0 & 80.0 & 35.0 & 99.5 & 97.2 & 100.0 & 13.6 & 86.7 & 78.0 & 54.0 & 78.3 \\
 & AdaKV & 100.0 & 100.0 & 95.0 & 100.0 & 90.0 & 61.0 & 99.2 & 97.0 & 100.0 & 13.7 & 92.0 & 79.0 & 55.0 & 83.2 \\
 & PyramidKV & 100.0 & 100.0 & 53.0 & 100.0 & 79.0 & 19.0 & 99.8 & 95.2 & 99.8 & 5.3 & 84.3 & 77.0 & 54.0 & 74.3 \\
 & AnchorKV & 100.0 & 100.0 & 99.0 & 100.0 & 99.0 & 95.0 & 99.8 & 96.2 & 100.0 & 21.4 & 94.0 & 82.0 & 57.0 & \textbf{88.0} \\
\midrule
\multirow{4}{*}{$20\times$} & SnapKV & 100.0 & 99.0 & 36.0 & 100.0 & 71.0 & 16.0 & 99.2 & 90.2 & 100.0 & 5.0 & 83.0 & 76.0 & 56.0 & 71.7 \\
 & AdaKV & 100.0 & 99.0 & 56.0 & 100.0 & 80.0 & 35.0 & 99.8 & 95.0 & 100.0 & 12.0 & 86.0 & 79.0 & 55.0 & 76.7 \\
 & PyramidKV & 100.0 & 99.0 & 25.0 & 100.0 & 69.0 & 6.0 & 99.5 & 90.8 & 99.6 & 3.1 & 82.7 & 78.0 & 56.0 & 69.9 \\
 & AnchorKV & 100.0 & 100.0 & 100.0 & 100.0 & 98.0 & 92.0 & 99.8 & 96.5 & 100.0 & 21.4 & 93.7 & 77.0 & 60.0 & \textbf{87.6} \\
\bottomrule
\end{tabular}
\caption{RULER at 64K on Llama-3.1-70B-Instruct.}
\label{tab:ruler64k-llama70b}
\end{table*}
\begin{table*}[t]
\centering
\setlength{\tabcolsep}{1.5pt}
\small
\begin{tabular}{@{}llrrrrrrrrrrrrrrrrr@{}}
\toprule
 &  & \multicolumn{3}{c}{Single-Doc QA} & \multicolumn{3}{c}{Multi-Doc QA} & \multicolumn{3}{c}{Summarization} & \multicolumn{3}{c}{Few-shot} & \multicolumn{2}{c}{Synthetic} & \multicolumn{2}{c}{Code} &  \\
\cmidrule(lr){3-5}\cmidrule(lr){6-8}\cmidrule(lr){9-11}\cmidrule(lr){12-14}\cmidrule(lr){15-16}\cmidrule(lr){17-18}
Ratio & Method & NQA & Qsp & MFQA & HQA & 2Wik & MSQ & GovR & QMS & MNws & TREC & TQA & SAMS & PCnt & PRe & LCC & RB-P & Avg \\
\midrule
$1\times$ & FullKV & 35.8 & 49.4 & 54.7 & 65.7 & 68.5 & 46.0 & 34.8 & 24.3 & 26.7 & 74.0 & 80.9 & 47.8 & 17.5 & 98.5 & 69.3 & 60.9 & 53.4 \\
\midrule
$4.2\times$ & TurboQuant & 35.2 & 48.6 & 54.6 & 63.0 & 69.7 & 48.2 & 35.3 & 23.7 & 26.9 & 74.0 & 80.9 & 47.9 & 18.5 & 98.5 & 69.0 & 60.4 & 53.4 \\
\midrule
\multirow{4}{*}{$10\times$} & SnapKV & 37.8 & 42.1 & 51.7 & 65.0 & 68.5 & 48.3 & 27.1 & 22.8 & 20.3 & 53.0 & 89.6 & 43.1 & 14.0 & 98.0 & 68.6 & 66.0 & 51.0 \\
 & AdaKV & 37.0 & 43.8 & 52.9 & 65.4 & 68.0 & 49.3 & 28.1 & 23.9 & 21.1 & 57.5 & 89.9 & 45.5 & 18.5 & 98.0 & 69.5 & 65.7 & 52.1 \\
 & PyramidKV & 37.0 & 42.2 & 51.4 & 64.0 & 68.2 & 47.4 & 26.6 & 23.1 & 20.1 & 53.0 & 87.3 & 42.5 & 14.5 & 98.0 & 68.5 & 65.6 & 50.6 \\
 & AnchorKV & 35.8 & 48.7 & 54.9 & 64.8 & 69.5 & 47.4 & 33.4 & 24.0 & 25.7 & 74.0 & 80.7 & 47.6 & 17.5 & 98.0 & 68.0 & 58.5 & \textbf{53.0} \\
\midrule
\multirow{4}{*}{$20\times$} & SnapKV & 35.2 & 34.8 & 45.3 & 62.4 & 64.1 & 46.0 & 23.5 & 22.5 & 14.8 & 41.0 & 87.2 & 41.4 & 15.0 & 98.5 & 64.4 & 65.5 & 47.6 \\
 & AdaKV & 36.6 & 37.8 & 50.8 & 65.1 & 65.1 & 49.1 & 24.9 & 22.7 & 15.4 & 48.0 & 93.0 & 42.9 & 17.0 & 98.0 & 64.9 & 66.9 & 49.9 \\
 & PyramidKV & 35.4 & 35.5 & 45.7 & 62.4 & 64.0 & 46.2 & 23.7 & 22.4 & 14.8 & 41.0 & 87.2 & 41.4 & 15.0 & 98.5 & 64.4 & 65.4 & 47.7 \\
 & AnchorKV & 36.2 & 49.0 & 53.4 & 64.6 & 67.8 & 47.7 & 30.2 & 23.3 & 22.3 & 75.0 & 80.8 & 46.9 & 17.5 & 98.5 & 69.1 & 59.5 & \textbf{52.6} \\
\bottomrule
\end{tabular}
\caption{LongBench on Llama-3.1-70B-Instruct.}
\label{tab:longbench-llama70b}
\end{table*}

\newpage
% \input{Other/ReproducibilityChecklist}

\end{document}